\documentclass{article}

\usepackage[preprint]{neurips_2024}

\usepackage[utf8]{inputenc} 
\usepackage[T1]{fontenc}    
\usepackage{hyperref}       
\usepackage{url}            
\usepackage{booktabs}       
\usepackage{amsfonts}       
\usepackage{nicefrac}       
\usepackage{xcolor}
\usepackage[nointegrals]{wasysym}
\usepackage{microtype}      
\usepackage{caption}
\usepackage{listings}
\usepackage[algo2e, ruled]{algorithm2e}
\SetKwComment{Comment}{$\triangleright$\ }{}
\SetKwProg{Init}{Initialize}{}{}
\usepackage{graphicx}
\usepackage{subfigure}
\usepackage{placeins}
\usepackage{subcaption}
\usepackage{bbm}
\usepackage{makecell}
\usepackage{multirow}
\usepackage{tikz}
\usepackage{amsmath}
\usepackage{amssymb}
\usepackage{mathtools}
\usepackage{amsthm}
\usepackage{nccmath}
\usepackage{algorithm}
\usepackage{wrapfig}
\usepackage{xspace}
\usepackage{tcolorbox}
\usepackage{colortbl}
\usepackage{fancyhdr,lipsum}
\usepackage[export]{adjustbox}
\usepackage{enumitem} 

\tcbuselibrary{most}

\definecolor{darkred}{RGB}{200, 0, 0}
\definecolor{darkblue}{RGB}{0, 0, 200}

\title{\textsc{FuseChat}: Knowledge Fusion of Chat Models}

\author{Fanqi Wan, Longguang Zhong, Ziyi Yang, Ruijun Chen, Xiaojun Quan\thanks{$\;\;$Corresponding authors.} \\
School of Computer Science and Engineering, Sun Yat-sen University, China \\
\texttt{wanfq@mail2.sysu.edu.cn, quanxj3@mail.sysu.edu.cn} \\ 
}

\begin{document}

\maketitle

\vspace{-0.60cm}
\begin{abstract}\label{abstract}
While training large language models (LLMs) from scratch can indeed lead to models with distinct capabilities and strengths, it incurs substantial costs and may lead to redundancy in competencies. Knowledge fusion aims to integrate existing LLMs of diverse architectures and capabilities into a more potent LLM through lightweight continual training, thereby reducing the need for costly LLM development. In this work, we propose a new framework for the knowledge fusion of chat LLMs through two main stages, resulting in \textsc{FuseChat}. Firstly, we conduct pairwise knowledge fusion on source chat LLMs of varying structures and scales to create multiple target LLMs with identical structure and size via lightweight fine-tuning. During this process, a statistics-based token alignment approach is introduced as the cornerstone for fusing LLMs with different structures. Secondly, we merge these target LLMs within the parameter space, where we propose a novel method for determining the merging coefficients based on the magnitude of parameter updates before and after fine-tuning. We implement and validate \textsc{FuseChat} using six prominent chat LLMs with diverse architectures and scales, including \texttt{OpenChat-3.5-7B}, \texttt{Starling-LM-7B-alpha}, \texttt{NH2-SOLAR-10.7B}, \texttt{InternLM2-Chat-20B}, \texttt{Mixtral-8x7B-Instruct}, and \texttt{Qwen-1.5-Chat-72B}. Experimental results on two instruction-following benchmarks, AlpacaEval 2.0 and MT-Bench, demonstrate the superiority of \texttt{\textsc{FuseChat-7B}} over baselines of various sizes. Our model is even comparable to the larger \texttt{Mixtral-8x7B-Instruct} and approaches \texttt{GPT-3.5-Turbo-1106} on MT-Bench as Figure \ref{fig:overall_result_mt_bench}. Our code, model weights, and data are public at \url{https://github.com/fanqiwan/FuseAI}.
\end{abstract}

\begin{figure*}[!ht] \centering
\vspace{-0.4cm}
\subfigure[] {
\label{fig:preliminary_res}     
\includegraphics[width=0.456\linewidth]{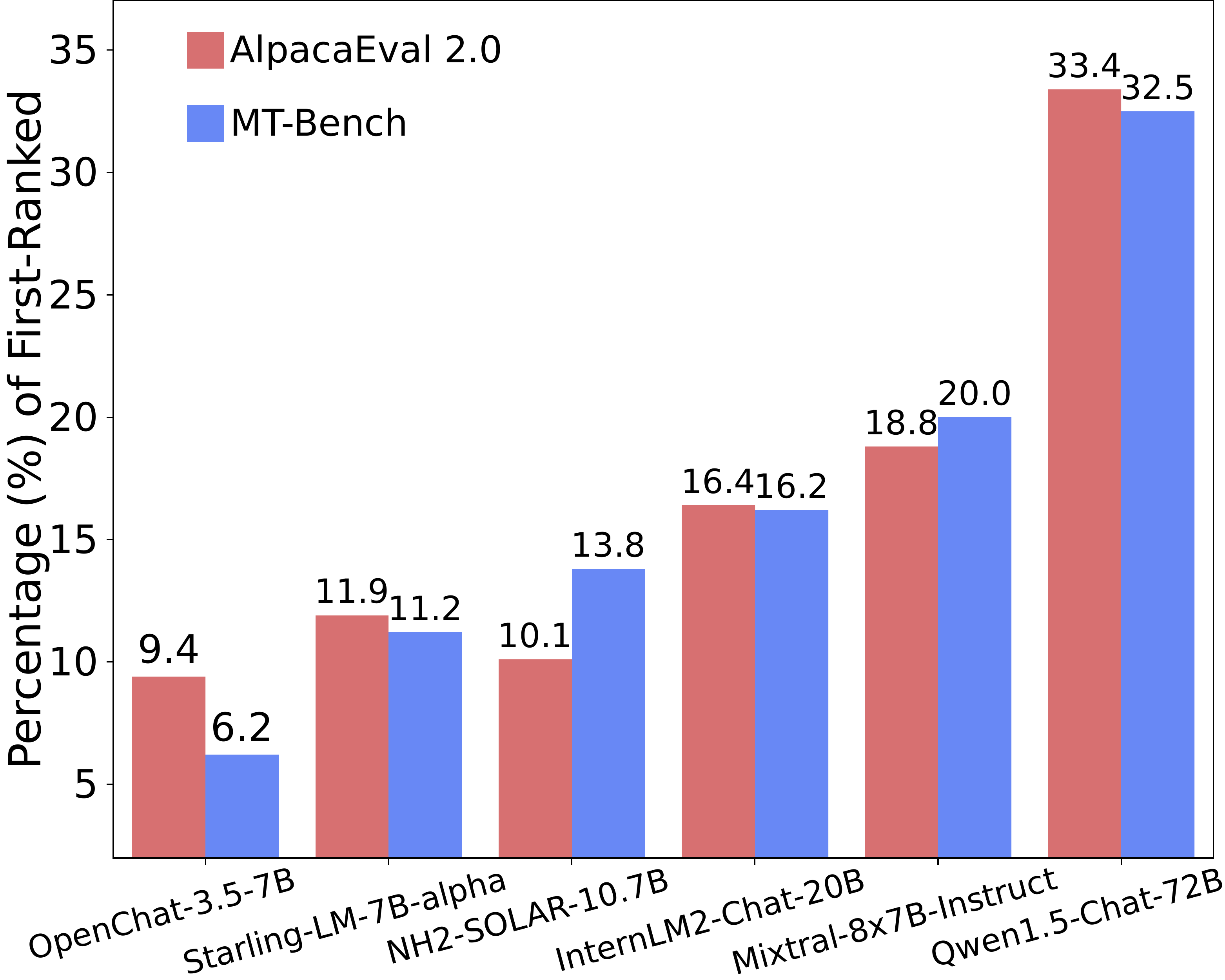}
}   
\hspace{0.02cm}
\subfigure[] {
\label{fig:overall_result_mt_bench}     
\includegraphics[width=0.456\linewidth]{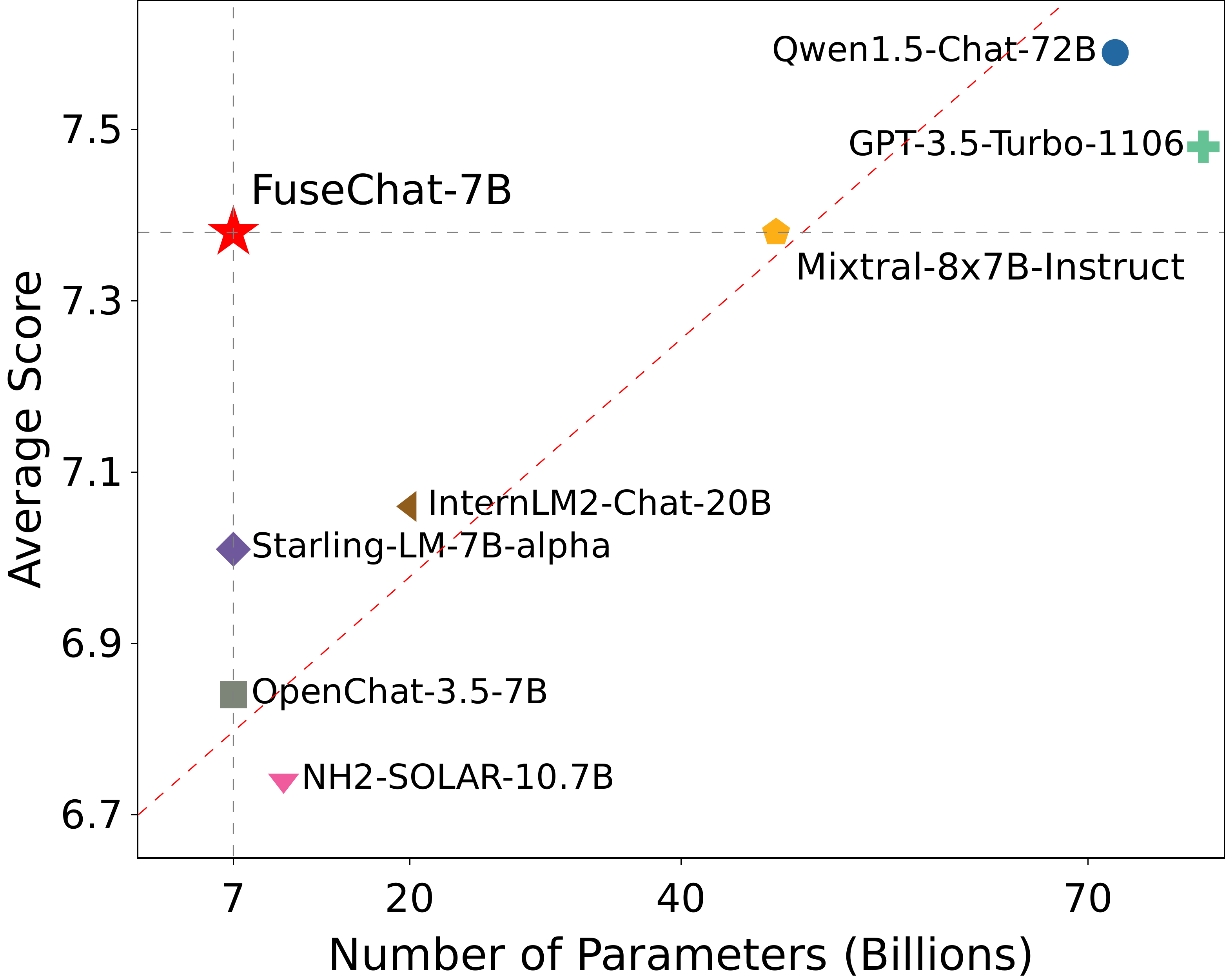}  
}     
\caption{Demonstration (left) of distinct strengths of existing chat LLMs and comparison (right) between \texttt{\textsc{FuseChat-7B}} and baseline LLMs. While the left figure plots the percentage of first-ranked responses of each LLM as measured by PairRM~\citep{jiang2023llm} on AlpacaEval 2.0 and MT-Bench, the right shows that \texttt{\textsc{FuseChat-7B}} achieves comparable performance to Mixtral-8x7B and approaches GPT-3.5 on MT-Bench. The red dashed line is linearly fitted from data points of all chat LLMs except \texttt{\textsc{FuseChat-7B}}.}     
\label{fig:abstract_fig} 
\vspace{-0.3cm}
\end{figure*}

\section{Introduction}\label{sec:introduction}
Large language models (LLMs) such as GPT~\citep{brown2020language} and LLaMA~\citep{touvron2023llama-2} series have demonstrated remarkable success across a wide range of natural language processing (NLP) tasks. 
Currently, it has become prevalent and imperative for individuals and corporations to build their own LLMs. However, the computational resources and time costs associated with LLM development remain prohibitively high. Furthermore, despite the structural and functional differences among LLMs, they often exhibit similar capabilities across various tasks. 
Therefore, besides training an LLM from scratch, another option is to combine the distinct advantages of existing LLMs into a more potent LLM, which is termed \emph{knowledge fusion of LLMs}~\citep{wan2024knowledge}. 
Figure \ref{fig:preliminary_res} illustrates the results of our preliminary experiment conducted on AlpacaEval 2.0 and MT-Bench, where we plot the percentage of questions each LLM answers best (measured by PairRM~\citep{jiang2023llm}) among six prominent chat LLMs. These established LLMs, regardless of their potency, exhibit distinct strengths. Therefore, knowledge fusion not only reduces the developmental costs of creating a new LLM but also has the potential to integrate the diverse strengths of existing models.

The endeavor to integrate the capabilities of multiple models has been a long-standing pursuit. 
For example, ensemble methods~\citep{littlestone1994weighted, jiang2023llm} directly aggregate the outputs of multiple models to enhance prediction performance and robustness. 
However, this approach requires maintaining multiple trained models during inference, which is inefficient for LLMs due to their substantial memory and inference time requirements. 
Another approach is to directly merge several neural networks into a single network through arithmetic operations in the parameter space~\citep{gupta2020stochastic}, whereas
this approach typically assumes uniform network architectures and requires manually-tuned~\citep{wortsman2022model, yadav2024ties} or automatically-learned~\citep{matena2022merging, jin2023dataless} coefficients to merge the parameters of different neural networks. 
In contrast, knowledge fusion~\citep{wan2024knowledge} seeks to integrate the capabilities of multiple LLMs, irrespective of their architectures, into a single LLM through lightweight continual training. This process essentially embodies a traditional multi-teacher knowledge distillation procedure~\citep{you2017learning}, but faces new challenges such as token alignment and fusion strategies across different LLMs.

In this study, we introduce a fuse-and-merge framework to extend the fusion of LLMs to chat-based LLMs\footnote{We will refer to ``chat-based LLMs'' simply as ``chat LLMs'' for brevity.} with diverse architectures and scales through two stages, resulting in \textsc{FuseChat}.
Firstly, we conduct pairwise knowledge fusion for source chat LLMs to generate multiple target LLMs of identical structure and size. To achieve this, we first select a pivot LLM and perform token alignment, followed by knowledge fusion between the pivot and each of the remaining LLMs. These target LLMs are expected to inherit the strengths of source chat LLMs through knowledge transfer during lightweight fine-tuning. 
Secondly, these target LLMs are merged within the parameter space, where we introduce a novel method called SCE to determine the merging coefficients based on the magnitude of parameter updates before and after fine-tuning. Moreover, SCE allocates parameter matrix-level coefficients that enable the merging at a fine-grained granularity without additional training efforts.

\textsc{FuseChat} offers superior potential compared to \textsc{FuseLLM}~\citep{wan2024knowledge}. 
Firstly, while \textsc{FuseLLM} limits its exploration to source LLMs of the same size as the target LLM, \textsc{FuseChat} lifts this restriction to enable the fusion of chat LLMs with varying sizes. 
This allows for greater adaptability to the fusion of diverse chat LLMs. Secondly, the framework of \textsc{FuseLLM} does not seamlessly support the inclusion of new source LLMs as it requires the combination of distribution matrices from all source LLMs during continual training. 
In contrast, integrating a new source LLM in \textsc{FuseChat} is plug-and-play, requiring only obtaining a target LLM from the new source LLM and merging it with the existing \textsc{FuseChat}. Thirdly, compared to many-to-one knowledge fusion, pairwise fusion empirically mitigates the challenges of knowledge distillation from source LLMs.

To verify the effectiveness of \textsc{FuseChat}, we implemented \texttt{\textsc{FuseChat-7B}} using six prominent open-source chat LLMs: \texttt{OpenChat-3.5-7B}~\citep{wang2024openchat}, \texttt{Starling-LM-7B-alpha}~\citep{zhu2023starling}, \texttt{NH2-SOLAR-10.7B}~\citep{kim2023solar}, \texttt{InternLM2-Chat-20B}~\citep{cai2024internlm2}, \texttt{Mixtral-8x7B-Instruct}~\citep{jiang2024mixtral}, and \texttt{Qwen-1.5-Chat-72B}~\citep{bai2023qwen}. Experimental results on two representative instruction-following benchmarks, AlpacaEval 2.0~\citep{dubois2024alpacafarm} and MT-Bench~\citep{zheng2024judging}, demonstrate the superiority of \texttt{\textsc{FuseChat-7B}} across a broad spectrum of chat LLMs at 7B, 10B, and 20B scales. Moreover, we validated the proposed token alignment method and the SCE merging method through a series of analytical experiments.

\vspace{-0.3cm}\section{Related Work}
\label{sec:related_work}
\vspace{-0.3cm}
\paragraph{Model Fusion} Combining the capabilities of diverse models has been a long-standing objective. Existing approaches to model fusion mainly fall into three categories. 
Firstly, traditional \emph{model ensemble} techniques combine the outputs of multiple models by weighted averaging~\citep{littlestone1994weighted} or majority voting~\citep{monteith2011turning} to enhance overall system performance. 
Recently, \cite{jiang2023llm} introduced a sequence-level ensemble framework for LLMs, which first conducts pairwise comparisons to rank the outputs of LLMs and then employs another LLM to consolidate the top-ranked candidates into an improved output.
In addition to the sequence-level ensemble, \cite{ding2024mastering} blended multiple LLMs using a token-level gating mechanism on the output logits.
To avoid additional training during ensemble, \cite{mavromatis2024pack} leveraged the perplexity of different LLMs over input prompts to determine the importance of each model.

Secondly, \emph{model merging} facilitates the fusion of models of identical structure and scale within the parameter space. \cite{wortsman2022model} combined multiple models, obtained by fine-tuning a model on the same dataset but with distinct strategies, through linear averaging.
\cite{matena2022merging} enhanced simple weighted average by incorporating Fisher Information Matrix~\citep{fisher1922mathematical} to determine the significance of individual model parameter.
\cite{jin2023dataless} performed merging by addressing an optimization problem that minimizes the L2 distance between merged and individual models, and conducting a closed-form solution. 
Although these methods can automatically compute merging coefficients, they necessitate either forward or backward propagation using additional data, making model merging compute-inefficient and memory-intensive.
\cite{ilharco2023editing} and \cite{zhang2023composing} conducted simple arithmetic operations on the task vectors or LoRA~\citep{hu2022lora} modules of different models, thereby enhancing multi-task ability and domain generalization.
To mitigate parameter interference, \cite{yu2023language} and \cite{yadav2024ties} introduced sparsification techniques that trim redundant values from task vectors prior to model merging.
Moreover, \cite{kim2023solar} and \cite{akiba2024evolutionary} advanced the field by merging multiple LLMs across both parameter and data flow spaces, yielding a merged LLM with up-scaled depth and superior performance.

Thirdly, \textit{mixture of experts} (MoEs) combines specialized expert modules with a sparsely activated mechanism~\citep{fedus2022switch}, presenting another venue for model fusion. 
\cite{komatsuzaki2023sparse} first proposed initializing a sparse MoEs module using multiple copies from a dense checkpoint.
To integrate multiple domain experts, \cite{sukhbaatar2024branch} trained multiple domain-specific LLMs from a seed LLM separately and then used feed-forward networks on top of these dense experts to instantiate a sparse MoEs module, followed by further fine-tuning to learn token-level routing. 
Similarly, \cite{feng2024mixture} trained multiple domain-specific LoRA~\citep{hu2022lora} modules as experts and combined these domain experts using an explicit sequence-level routing strategy.

Lastly, \textsc{FuseLLM}~\citep{wan2024knowledge} introduces another paradigm for the fusion of LLMs with structural differences. This approach builds upon knowledge distillation and leverages the probabilistic distribution matrices generated by source LLMs to transfer collective knowledge into a target LLM. Unlike model ensembles and MoEs, knowledge fusion does not require the parallel deployment of multiple models (experts). Furthermore, compared to model merging, which only applies to models with identical architectures, \textsc{FuseLLM} allows for the fusion of LLMs with different architectures.

\vspace{-0.4cm}
\paragraph{Knowledge Distillation} Knowledge fusion essentially performs knowledge distillation to transfer knowledge from source LLMs to a target LLM. Knowledge distillation \citep{hinton2015distilling} aims to train a small student model guided by one or more larger teacher models. Previous studies primarily focus on training a student model to mimic the teacher's behavior in text classification tasks, by replicating the teacher's output logits \citep{sanh2019distilbert, turc2019well}, as well as hidden states \citep{sun2019patient, jiao2020tinybert} and relations \citep{wang2020minilm}. In the realm of generative models, prevailing approaches maximize the log-likelihood of the student on the distributions \citep{khanuja2021mergedistill,gu2024minillm, agarwal2024onpolicy} or sequences \citep{kim2016sequence,peng2023instruction} generated by the teacher model. This paradigm can be extended to accommodate multiple teachers by either averaging the distributions \citep{you2017learning} or blending the sequences \citep{wang2024openchat}.

Compared to vanilla knowledge distillation, knowledge fusion of LLMs faces new challenges. Firstly, due to the differences in tokenization among various LLMs, token alignment is essential for transferring knowledge from source to target LLMs. Secondly, when dealing with distributions generated from multiple source LLMs, the fusion function becomes crucial for optimally integrating their distributions. Thirdly, to leverage the unique advantages of different LLMs, it is necessary and challenging to create a compact knowledge fusion dataset that is diverse in capabilities and domains.

\section{\textsc{FuseChat}}

\subsection{Overview}

\begin{figure*}[t]
    \centering
    \includegraphics[width=0.99\textwidth]{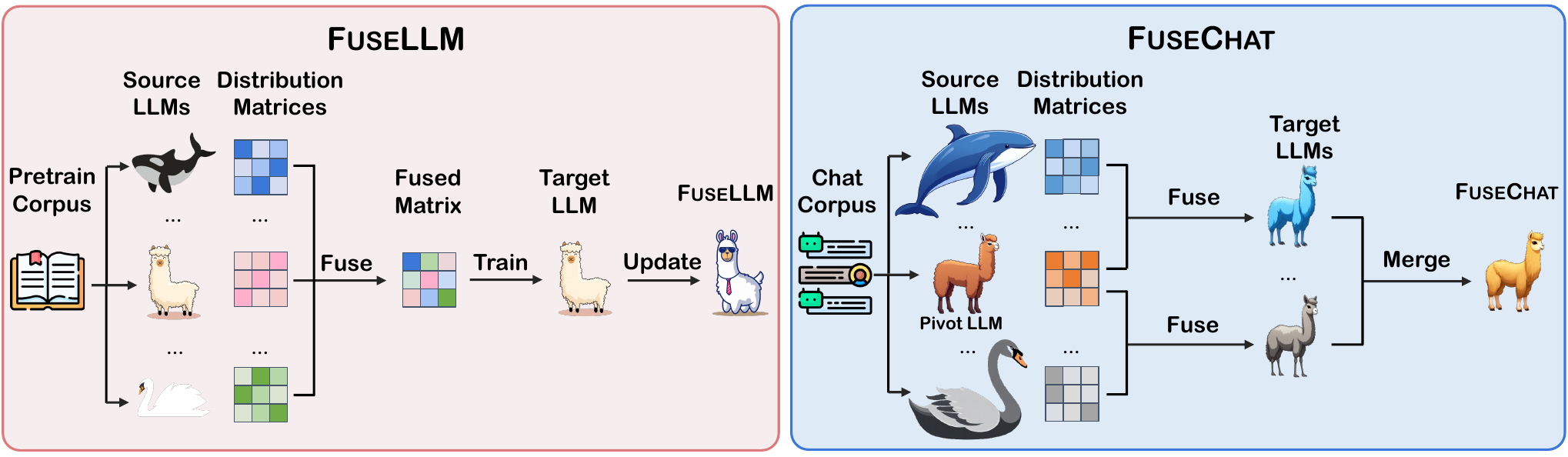}
	\caption{Overview of \textsc{FuseChat} in comparison with \textsc{FuseLLM}~\citep{wan2024knowledge}. Distinct animal icons symbolize different LLMs, where each species and size indicate a unique architecture and scale, respectively.}
	\label{fig:main}
	\vspace{-0.3cm}
\end{figure*}

Figure \ref{fig:main} presents an overview of our \textsc{FuseChat} in comparison with \textsc{FuseLLM}~\citep{wan2024knowledge}. The \textsc{FuseChat} framework consists of two main stages: fuse and merge. In the \emph{fuse} stage, pairwise knowledge fusion is conducted on source chat LLMs\footnote{We will use ``source chat LLMs'' and ``source LLMs'' interchangeably when there is no ambiguity.} to derive multiple target LLMs with identical structure and size. This process begins by selecting a pivot LLM, followed by performing knowledge fusion between the pivot and each remaining LLM. In the \emph{merge} stage, these target LLMs are combined within the parameter space, where we introduce a novel method to determine the merging coefficients based on the magnitude of parameter updates before and after fine-tuning.

Specifically, considering $K$ source LLMs \( \{ \mathcal{M}^s_i\}_{i=1}^{K} \) with varying architectures and scales, \textsc{FuseChat} first specifies one of the source LLMs, $\mathcal{M}^s_v$, as the pivot and then applies pairwise knowledge fusion to obtain ($K-1$) target LLMs, \( \{ \mathcal{M}^t_j\}_{j=1}^{K-1} \), which share the same architecture and initialized weights as the pivot LLM. The selection of the pivot depends on the desired structure and scale for the target LLMs, while also considering the capabilities and performance of a candidate LLM. 

To perform pairwise knowledge fusion, \textsc{FuseChat} prompts these source LLMs using a supervised fine-tuning dataset $\mathcal{D} = \{I_i, R_i\}_{i=1}^{M}$ to showcase their inherent knowledge by responding to each instruction in $\mathcal{D}$. Token alignment~\citep{fu2023specializing,wan2024knowledge} between the source LLMs and the pivot is then conducted to properly map the resulting probabilistic distribution matrices. These distribution matrices are subsequently used for pairwise knowledge fusion~\citep{wan2024knowledge} through lightweight fine-tuning to obtain ($K-1$) target LLMs. 
Following this, the target LLMs are merged in the parameter space to yield the final fused LLM \(\mathcal{M}^f\). 
To incorporate fine-grained advantages of target LLMs, we introduce a new merging method named SCE to obtain the merging coefficients based on \emph{selection}, \emph{calculation}, and \emph{erasure} on the task vectors~\citep{ilharco2023editing} which represent variation of model weights before and after fine-tuning. SCE enables the automatic allocation of parameter matrix-level merging coefficients, facilitating the merging of LLMs at a finer granularity.

\subsection{Preliminaries}

Given an instruction $I_i$ and the corresponding response $R_i$ of length $N$ from the fine-tuning dataset $\mathcal{D}$, we use $R_{i, <t} = (r_{i, 1}, r_{i, 2}, \ldots, r_{i, t-1})$ to represent the sequence preceding the $t$th token in the response. 
The supervised fine-tuning (SFT) objective for a pre-trained language model parameterized by $\theta$ is defined as minimizing the following negative log-likelihood:
\begin{equation}\label{eqn:token_level_clm}
\mathcal{L}_{\text{SFT}} = -\mathbb{E}_{(I_{i}, R_{i}) \sim \mathcal{D}}\left[\sum_{t\leq N} \log p_{\theta}(r_{i, t} | R_{i, <t}, I_{i})\right],
\end{equation}
where $p_{\theta}(r_{i, t} | R_{i, <t}, I_{i})$ is the model's predicted probability for the $t$th token $r_{i,t}$ in $R_i$ given the instruction and preceding tokens in the response.

\subsection{Pairwise Knowledge Fusion}
\label{sec:knowledge_fusion}

To facilitate the description of pairwise knowledge fusion, we reframe the above token-level view into a matrix format. Specifically, for each instruction $I_i$, we transform the token-level predictions into a probabilistic distribution matrix, $\mathbf{P}_i^\theta\in \mathbb{R}^{N \times V}$, where $V$ denotes the vocabulary size. The distribution matrix is assumed to reflect certain inherent knowledge of the language model in responding to the instruction \citep{wan2024knowledge}. Consequently, different probabilistic distribution matrices obtained from different chat LLMs can be used to represent the diverse knowledge embedded within these models. Based on this assumption, \textsc{FuseChat} performs pairwise knowledge fusion by fine-tuning the target LLMs, initialized from the pivot, using the probabilistic distribution matrices.

\paragraph{Model Fusion} 

For each instruction $I_i$ in $\mathcal{D}$, we first feed it into the $K$ source chat LLMs to obtain a set of probabilistic distribution matrices, denoted as $\{\mathbf{P}_i^{\theta_j}\}_{j=1}^{K}$, where $\theta_j$ represents the parameters of the $j$th chat LLM. 
Since these LLMs may employ different tokenizers, token alignment is necessary to properly map their probabilistic distribution matrices~\citep{fu2023specializing,wan2024knowledge}.
Then, pairwise knowledge fusion is conducted between the pivot LLM and each of the remaining source LLMs. To achieve this, we denote the probabilistic distribution matrix generated by the pivot LLM as $\mathbf{P}_i^{\theta_v}$ and merge it with each $\mathbf{P}_i^{\theta_j}|_{j\neq v}$ to obtain a set $\{\mathbf{P}_i^j\}_{j=1}^{K-1}$ of fused matrices as follows:
\begin{equation}\label{eqn:matrix_fusion}
\mathbf{P}_i^j = \mathbb{F}\text{usion}(\mathbf{P}_i^{\theta_v}, \mathbf{P}_i^{\theta_j})|_{j\neq v},
\end{equation}
where $\mathbb{F}\text{usion}(\cdot)$ represents the fusion function that merges two matrices. The resulting matrix $\mathbf{P}_i^j$ is seen as a representation of the collective knowledge and distinctive strengths of the two source LLMs. Among various fusion strategies, this work employs minimum cross-entropy (MinCE) \citep{wan2024knowledge} as the fusion function, which empirically performs the best. 

After that, we enforce alignment between the prediction of each target LLM \(\mathcal{M}^t_j\) and the corresponding fused representation matrices $\mathbf{P}_i^{j}$. We use $\mathbf{Q}_i^{\phi_j}$ to represent the output distribution matrix of target LLM $\mathcal{M}^t_j$ for instruction $I_i$ and define the fusion objective for training each target LLM as follows:

\begin{equation}\label{eqn:final_fusion_objective}
\mathcal{L}_{\text{Fusion}} = -\mathbb{E}_{(I_i, R_i) \sim \mathcal{D}}\left[\mathbb{H}(\mathbf{P}_i^{j}||\mathbf{Q}_i^{\phi_j})\right],
\end{equation}
where $\mathbb{H}(\cdot || \cdot)$ represents the cross entropy between two probabilistic distribution matrices.

The overall training objective for each pairwise knowledge fusion consists of a weighted combination of the supervised fine-tuning objective $ \mathcal{L}_{\text{SFT}} $ and the fusion objective $ \mathcal{L}_{\text{Fusion}}$ as follows:
\begin{equation}\label{eqn:final_objective}
\mathcal{L} = \lambda\mathcal{L}_{\text{SFT}} + (1-\lambda)\mathcal{L}_{\text{Fusion}}.
\end{equation}

\paragraph{Token Alignment} Token alignment aims to address the mappings of probabilistic distribution matrices $\{\mathbf{P}_i^{\theta_j}\in \mathbb{R}^{N \times V}\}_{j=1}^{K} $ generated by different source LLMs for a given instruction $I_i$. Therefore, the alignment involves two dimensions of the matrices: the sequence dimension for the tokenized response and the distribution dimension for the probabilistic distributions. In the sequence dimension, we follow previous works~\citep{fu2023specializing, wan2024knowledge} and adopt dynamic programming to recursively minimize the total cost of editing the tokens from a source LLM to align them with the pivot LLM. This process may result in 1-1, 1-n, and n-1 mappings, as shown in Figure \ref{fig:token_alignment_process}. In the distribution dimension, \cite{fu2023specializing} focused on aligning distributions based on the exact match (EM) between tokens in source and target distributions, which restricts the alignment to only 1-1 mappings and may result in too many unmatched tokens. \cite{wan2024knowledge} relaxed the EM constraint by aligning the distributions based on the minimum edit distance (MinED) between tokens in the vocabularies of source and target LLMs. While this approach improves the mapping success rate and expands to 1-n mappings, it ignores n-1 mappings and may introduce many misalignments.

In this work, we propose an enhanced token alignment strategy that utilizes mapping statistics (MS) from the sequence dimension as the criteria for alignment in the distribution dimension. 
We construct a global statistical matrix, where each column represents the frequency of mappings from a pivot token to all potential source tokens, derived from sequence-dimensional token alignments. 
In the case of 1-1 and 1-n mappings, we align the distributions based on the maximum mapping frequency in the respective columns of the statistical matrix for each pivot token in the distribution. 
For n-1 mappings, we first calculate a weighted average of the source tokens' distributions according to their mapping frequencies in the statistical matrix to obtain a merged distribution. This merged distribution is then aligned to the pivot distribution similar to the procedure employed for 1-1 mappings. As illustrated in Figure \ref{fig:token_alignment_process}, this approach better reflects the token mapping statistics in the dataset, thereby preserving significant information in the aligned distribution matrices while minimizing alignment errors.

\subsection{Model Merging}
\label{sec:model_merge}

Since the target LLMs \( \{ \mathcal{M}^t_j\}_{j=1}^{K-1} \) resulting from pairwise knowledge fusion share identical architecture and scale while possessing diverse advantages and capabilities learned from the source LLMs, we further merge them in the parameter space to obtain the final fused LLM $\mathcal{M}^f$.
To ensure the adaptability and scalability of \textsc{FuseChat}, it is essential to maintain the simplicity of the merging strategy. 
Primarily, the calculation of merging coefficients should be automated, obviating the complex hyperparameter tuning. Secondly, the merging procedure should not demand forward or backward propagation over additional data, which is computationally inefficient and memory-intensive.

As described in Algorithm \ref{alg:sce}, we propose a novel merging method named SCE (\emph{select}, \emph{calculate}, and \emph{erase}) for parameter matrix-level merging. Analogous to task vectors~\citep{ilharco2023editing}, we first define fusion vectors $\{\delta_{j}\}_{j=1}^{K-1}$ (Eq. \ref{eqn:variation}) as the direction and magnitude of weight updates from pivot LLM $\mathcal{M}^s_v$ to target LLMs $\{\mathcal{M}^t_j\}_{j=1}^{K-1}$ during model fusion. For each parameter matrix unit in target LLMs, we derive the merged weights using fusion vectors through a three-step process.

\begin{wrapfigure}{r}{0.50\textwidth}
\vspace{-0.75cm}
\begin{minipage}{0.99\linewidth}
\begin{algorithm}[H]
\caption{SCE Procedure}
\label{alg:sce}
\DontPrintSemicolon
\KwIn{target LLMs parameters \(\{\phi_{j}\}_{j=1}^{K-1}\), pivot LLM parameters \(\theta_{v}\), threshold $\tau$.}
\KwOut{merged LLM parameters \(\Phi\)}

\Comment{Create fusion vectors}\vspace{-1.1mm}
\begin{equation}\label{eqn:variation}
\{\delta_{j}\}_{j=1}^{K-1} = \{\phi_{j} - \theta_{v}\}_{j=1}^{K-1}
\end{equation}\vspace{-4.2mm}

\Comment{Calculate parameter matrix-level merging coefficients}\vspace{1.1mm}
\For{\(\{\delta_{j,m}\}_{j=1}^{K-1} \in \{\delta_{j}\}_{j=1}^{K-1}\)}{
\vspace{1.1mm}
    \Comment{Step 1: Select salient elements}\vspace{-1.5mm}
    \begin{equation}\label{eqn:select}
    \{{\hat{\delta}}_{j,m}\}_{j=1}^{K-1} = \text{Select}(\{\delta_{j,m}\}_{j=1}^{K-1},\tau)
    \end{equation}\vspace{-4.5mm}
    
    \Comment{Step 2: Calculate coefficients}\vspace{-1.5mm}
    \begin{equation}\label{eqn:calculate}
    \{\eta_{j,m}\}_{j=1}^{K-1} = \text{Calculate}(\{{\hat{\delta}}_{j,m}^2\}_{j=1}^{K-1})
    \end{equation}\vspace{-4.2mm}

    \Comment{Step 3: Erase minority elements}\vspace{-1.5mm}
    \begin{equation}\label{eqn:erase}
    \{\delta^{'}_{j,m}\}_{j=1}^{K-1} = \text{Erase}(\{{\hat{\delta}}_{j,m}\}_{j=1}^{K-1})
    \end{equation}\vspace{-4.2mm}

    \Comment{Update merged LLM parameters}\vspace{-1.5mm}
    \begin{equation}\label{eqn:merge}
    \Phi_m = \theta_{v,m} + \sum_{j=1}^{K - 1} \eta_{j,m} \delta^{'}_{j,m}
    \end{equation}\vspace{-4.5mm}
}
\Return{\(\Phi\)}
\end{algorithm}
\end{minipage}
\vspace{-1.8cm}
\end{wrapfigure}

(1) \textbf{Select}: During the pairwise knowledge fusion, target LLMs dynamically evolve their parameters to incorporate the advantages of their corresponding source LLMs. Fusion vectors for each parameter matrix unit with substantial variations across different target LLMs are supposed to signify distinctive and significant strengths. Therefore, we first select the top $\tau\%$ elements from each parameter matrix-level fusion vector $\{\delta_{j,m}\}_{j=1}^{K-1}$ with high variance across multiple target LLMs, resulting in $\{\hat\delta_{j,m}\}_{j=1}^{K-1}$ (Eq. \ref{eqn:select}).
(2) \textbf{Calculate}: We then calculate the sum of squares of elements in $\hat\delta_{j,m}$ and obtain a matrix-level merging coefficient for each target LLM as $\eta_{j,m} = \frac{\sum\hat\delta^2_{j,m}}{\sum_j\sum\hat\delta^2_{j,m}}$.
(3) \textbf{Erase}: Each parameter may exhibit conflicting signs across fusion vectors from different target LLMs, which could cause interference during model merging~\citep{yadav2024ties}. Thus, for each parameter we sum its values in $\{\hat\delta_{j,m}\}_{j=1}^{K-1}$ across target LLMs and erase elements with minority directions (Eq. \ref{eqn:erase}). Finally, the filtered $\{\delta^{'}_{j,m}\}_{j=1}^{K-1}$ are merged based on the calculated coefficients, and added to the pivot LLM's parameters (Eq. \ref{eqn:merge}).

\subsection{Discussions}
The reasons why \textsc{FuseChat} adopts pairwise rather than many-to-one knowledge fusion as \textsc{FuseLLM}~\citep{wan2024knowledge} are twofold. Firstly, directly fusing all the source LLMs proves to be difficult, as evidenced by the results of \texttt{OpenChat-3.5-7B Multi} in Table \ref{tab:main_res}. Instead, \textsc{FuseChat} adopts a fuse-and-merge strategy, wherein the fusing stage employs pairwise knowledge fusion between the pivot LLM and other source LLMs, which reduces the difficulty of model fusion. Secondly, \textsc{FuseChat} offers superior scalability compared to \textsc{FuseLLM}. The framework of \textsc{FuseLLM} requires the combination of distribution matrices from all source LLMs during continual training, which does not easily support the inclusion of new LLMs. In contrast, \textsc{FuseChat} supports plug-and-play integration of new source LLMs at any scale. This requires only obtaining a target LLM by fusing the new source LLM with the pivot, and then merging it with the existing version of \textsc{FuseChat}.

\vspace{-0.2cm}
\section{Experiments}
\vspace{-0.2cm}

In our experiments, we explore the fusion of chat LLMs with diverse architectures and scales.~Specifically, we conduct experiments using six representative chat LLMs as the source LLMs, including \texttt{OpenChat-3.5-7B}~\citep{wang2024openchat}, \texttt{Starling-LM-7B-alpha}~\citep{zhu2023starling}, \texttt{NH2-SOLAR-10.7B}~\citep{kim2023solar}, \texttt{InternLM2-Chat-20B}~\citep{cai2024internlm2}, \texttt{Mixtral-8x7B-Instruct}~\citep{jiang2024mixtral}, and \texttt{Qwen-1.5-Chat-72B}~\citep{bai2023qwen}. 
As for the pivot LLM, which also serves as the starting point for the target LLMs, we opt for \texttt{OpenChat-3.5-7B} due to its balanced scale and performance. To begin, we first apply pairwise knowledge fusion (Section \ref{sec:knowledge_fusion}) to create five distinct target LLMs with the same structure. These target LLMs are then merged using the SCE method (Section \ref{sec:model_merge}), resulting in the final \texttt{\textsc{FuseChat-7B}}.

\vspace{-0.15cm}
\subsection{Experimental Setup}
\label{sec:experimental_setup}
\vspace{-0.1cm}
\paragraph{Training Dataset}
To leverage the strengths of source LLMs during knowledge fusion while alleviating catastrophic forgetting, we curate a high-quality dataset named \textsc{FuseChat-Mixture} from two different sources. First, 50\% of the training instances are sampled from the dataset used by the pivot LLM, \texttt{OpenChat-3.5-7B}. Second, we gather the remaining training instances, which have not been encountered by the pivot LLM, from open-source communities.  These two sources result in a corpus comprising approximately  95,000 dialogues across spanning various domains. For further details on \textsc{FuseChat-Mixture}, please refer to Appendix \ref{appendix: details_of_training_dataset}.
\vspace{-0.2cm}
\paragraph{Training Details}
In all experiments, we train the target LLMs using a batch size of 128 and a maximum length of 2048 on a single node with 8x80GB NVIDIA A800 GPUs for three epochs, which takes approximately 9 hours. 
The models are optimized using the AdamW~\citep{loshchilov2019decoupled} optimizer with $\beta_{1}=0.9$ and $\beta_{2}=0.999$.
We use a weight decay of 0.0 and gradient clipping of 1.0. 
A cosine learning rate schedule is employed, with a maximum learning rate of 5e-6 and a warmup ratio of 0.03.
We empirically set the combination weight $\lambda$ in Eq. \ref{eqn:final_objective} to 0.9. 
Our training framework is implemented based on the HuggingFace Transformers~\citep{wolf2020transformers}. 
\vspace{-0.2cm}
\paragraph{Evaluation}
We assess the performance of \texttt{\textsc{FuseChat-7B}} on two representative benchmarks to evaluate its ability to follow instructions and engage in conversations effectively. The first benchmark, AlpacaEval 2.0~\citep{dubois2024alpacafarm}, comprises 805 instructions across five test subsets. It compares the Win Rate and Length-Controlled Win Rate (LC Win Rate)~\citep{dubois2024length} of a model against GPT-4. We employ the default settings and utilize GPT-4 (\texttt{GPT-4-1106-Preview}) to evaluate the quality of generated responses. The second benchmark, MT-Bench~\citep{zheng2024judging}, consists of 80 multi-turn dialogues spanning various domains including writing, roleplay, reasoning, math, coding, STEM, and humanities. Originally, GPT-4 (\texttt{GPT-4-0613}) was used as the evaluator, providing a scalar score ranging from 1 (lowest) to 10 (highest) for each generated response. However, due to inaccuracies in the reference responses, we adopt an updated version, \texttt{GPT-4-0125-Preview}, as per the latest works\footnote{\url{https://github.com/lm-sys/FastChat/pull/3158}}, to correct the errors and evaluate the generated responses.
\vspace{-0.2cm}
\paragraph{Baselines}
In our experiments, we compare our \texttt{\textsc{FuseChat-7B}} with four categories of baseline LLMs, including (i) Proprietary LLMs, (ii) Source LLMs, (iii) Ensemble LLMs, and (iv) Fused LLMs. The details of these baselines are shown in Appendix \ref{appendix: details_of_baselines}.
\vspace{-0.15cm}
\subsection{Overall Results}
\label{sec:overall_results}
\vspace{-0.1cm}
In Table \ref{tab:main_res}, we present the overall results of \texttt{\textsc{FuseChat-7B}} in comparison with baselines of various architectures and scales on AlpacaEval 2.0 and MT-Bench. 
Our key observations are as follows. Firstly, after supervised fine-tuning on our high-quality dataset, \texttt{OpenChat-3.5-7B SFT} demonstrates slightly better performance than the pivot LLM \texttt{OpenChat-3.5-7B}. 
Secondly, in comparison to \texttt{Openchat-3.5-7B Multi}, which fuses multiple source LLMs simultaneously as \textsc{FuseLLM}~\citep{wan2024knowledge}, the target LLMs resulting from pairwise knowledge fusion exhibit superior performance, demonstrating the effectiveness of pairwise fusion in reducing the fusion difficulty. For instance, through the integration of \texttt{OpenChat-3.5-7B} and \texttt{Mixtral-8x7B-Instruct}, the fused target LLM \texttt{OpenChat-3.5-7B Mixtral} achieves relative gains of 13.93\% LC Win Rate and 5.23\% Average Score over \texttt{OpenChat-3.5-7B SFT}, significantly surpassing \texttt{OpenChat-3.5-7B Multi}.
Furthermore, after merging these target LLMs, \texttt{\textsc{FuseChat-7B}} shows substantial performance enhancements of 18.34\% and 7.27\% in the two metrics. 
This illustrates the superiority of \texttt{\textsc{FuseChat-7B}} across a range of source LLMs at various scales, even comparable to 8x7B MoEs and approaching GPT-3.5. 

Moreover, in comparison to the ensemble LLMs of 162B, which generate the 1st response from six parallel deployed LLMs based on different ranking criteria, \texttt{\textsc{FuseChat-7B}} outperforms most of these LLMs except \texttt{Top1-GPT4} on MT-Bench, while being 23x smaller and independent of GPT-4.

\begin{table*}[!t]
\centering
\resizebox{0.999\linewidth}{!}{
    \begin{tabular}{llccccc}
    \toprule
        \multirow{3}{*}{\textbf{Model}} & \multirow{3}{*}{\textbf{\#Params}} & \multicolumn{2}{c}{\textbf{AlpacaEval 2.0}} & \multicolumn{3}{c}{\textbf{MT-Bench}} \\ 
        ~ & ~ & \multicolumn{2}{c}{\textbf{(GPT-4-1106-Preview)}} & \multicolumn{3}{c}{\textbf{(GPT-4-0125-Preview)}} \\ \cmidrule(lr){3-4} \cmidrule(lr){5-7}
        ~ & ~ & \textbf{Win Rate} & \textbf{LC Win Rate} & \textbf{1st Turn} & \textbf{2nd Turn} & \textbf{Average Score} \\
        \midrule
        \multicolumn{7}{c}{\textbf{Proprietary LLMs}} \\ \midrule
        GPT-3.5-Turbo-1106\citep{achiam2023gpt} & - & 9.18 & 19.30 & 7.56 & 7.41 & 7.48 \\ 
        Claude-3-Opus\citep{anthropic2024claude} & - & 29.04 & 40.39 & 8.84 & 8.30 & 8.57 \\ 
        GPT-4-1106-Preview\citep{achiam2023gpt} & - & 50.00 & 50.00 & 8.86 & 8.71 & 8.79 \\
        \midrule
        \multicolumn{7}{c}{\textbf{Source LLMs}} \\ \midrule
        OpenChat-3.5-7B\citep{wang2024openchat} & 7B  & 10.20 & 14.90 & 7.14 & 6.55 & 6.84 \\ 
        Starling-LM-7B-alpha\citep{zhu2023starling} & 7B  & 14.20 & 14.70 & 7.54 & 6.49 & 7.01 \\ 
        NH2-SOLAR-10.7B\citep{kim2023solar} & 10.7B  & 12.22 & 18.13 & 7.11 & 6.36 & 6.74 \\
        InternLM2-Chat-20B\citep{cai2024internlm2} & 20B  & 21.70 & 18.70 & 7.78 & 6.34 & 7.06 \\ 
        Mixtral-8x7B-Instruct\citep{jiang2024mixtral} & 8x7B  & 18.30 & 23.70 & 7.76 & 7.00 & 7.38 \\ 
        Qwen1.5-Chat-72B\citep{bai2023qwen} & 72B  & 26.50 & 36.60 & 7.83 & 7.36 & 7.59 \\ \midrule
        \multicolumn{7}{c}{\textbf{Ensemble LLMs}} \\ \midrule
        Top1-PPL\citep{mavromatis2024pack} & 162B  & 25.11 & 27.97 & 7.79 & 6.95 & 7.37 \\
        Top1-LLM-Blender\citep{jiang2023llm} & 162B  & 24.45 & 29.11 & 7.85 & 6.70 & 7.28 \\
        Top1-GPT4\citep{achiam2023gpt} & 162B  & 42.82 & 43.87 & 8.79 & 8.01 & 8.40 \\ \midrule
        \multicolumn{7}{c}{\textbf{Fused LLMs}} \\ \midrule
        OpenChat-3.5-7B SFT & 7B  & 10.56 & 14.50 & 7.36 & 6.40 & 6.88 \\ 
        OpenChat-3.5-7B Multi & 7B & 10.19 (\textcolor{darkred}{-3.50\%}) & 13.43 (\textcolor{darkred}{-7.38\%}) & \underline{7.69} (\textcolor{darkblue}{+4.48\%}) & 6.26 (\textcolor{darkred}{-2.19\%}) & 6.99 (\textcolor{darkblue}{+1.60\%}) \\ 
        OpenChat-3.5-7B Starling & 7B  & 11.43 (\textcolor{darkblue}{+8.24\%}) & 16.20 (\textcolor{darkblue}{+11.72\%}) & \underline{7.69} (\textcolor{darkblue}{+4.48\%}) & 6.73 (\textcolor{darkblue}{+5.16\%}) & 7.22 (\textcolor{darkblue}{+4.94\%}) \\ 
        OpenChat-3.5-7B SOLAR & 7B  & 11.12 (\textcolor{darkblue}{+5.30\%}) & 16.51 (\textcolor{darkblue}{+13.86\%}) & 7.58 (\textcolor{darkblue}{2.99\%}) & 6.76 (\textcolor{darkblue}{+5.62\%}) & 7.17 (\textcolor{darkblue}{+4.22\%})\\ 
        OpenChat-3.5-7B InternLM & 7B & \textbf{11.82} (\textcolor{darkblue}{+11.93\%}) & 15.21 (\textcolor{darkblue}{+4.90\%}) & 7.63 (\textcolor{darkblue}{+3.67\%}) & 6.78 (\textcolor{darkblue}{+5.94\%}) & 7.21 (\textcolor{darkblue}{+4.80\%}) \\
        OpenChat-3.5-7B Mixtral & 7B & \underline{11.74} (\textcolor{darkblue}{+11.17\%}) & \underline{16.52} (\textcolor{darkblue}{+13.93\%}) & 7.58 (\textcolor{darkblue}{+2.99\%}) & \underline{6.90} (\textcolor{darkblue}{+7.81\%}) & \underline{7.24} (\textcolor{darkblue}{+5.23\%}) \\ 
        OpenChat-3.5-7B Qwen & 7B & 10.93 (\textcolor{darkblue}{+3.50\%}) & 14.98 (\textcolor{darkblue}{+3.31\%}) & \underline{7.69} (\textcolor{darkblue}{+4.48\%}) & 6.78 (\textcolor{darkblue}{+5.94\%}) & 7.23 (\textcolor{darkblue}{+5.09\%}) \\ 
        \midrule
        \textsc{FuseChat-7B} & 7B & 11.52 (\textcolor{darkblue}{+9.09\%}) & \textbf{17.16} (\textcolor{darkblue}{+18.34\%}) & \textbf{7.70} (\textcolor{darkblue}{+4.62\%}) & \textbf{7.05} (\textcolor{darkblue}{+10.16\%}) & \textbf{7.38} (\textcolor{darkblue}{+7.27\%})\\ \bottomrule 
    \end{tabular}
}
\caption{Overall results of the proposed \texttt{\textsc{FuseChat-7B}} and baselines on AlpacaEval 2.0 and MT-Bench. Percentages indicate the rate of improvement/decrease compared to \texttt{OpenChat-3.5-7B SFT}.}
\label{tab:main_res}
\vspace{-0.3cm}
\end{table*}

\begin{figure*}[!t]
    \centering
    \includegraphics[width=0.999\textwidth]{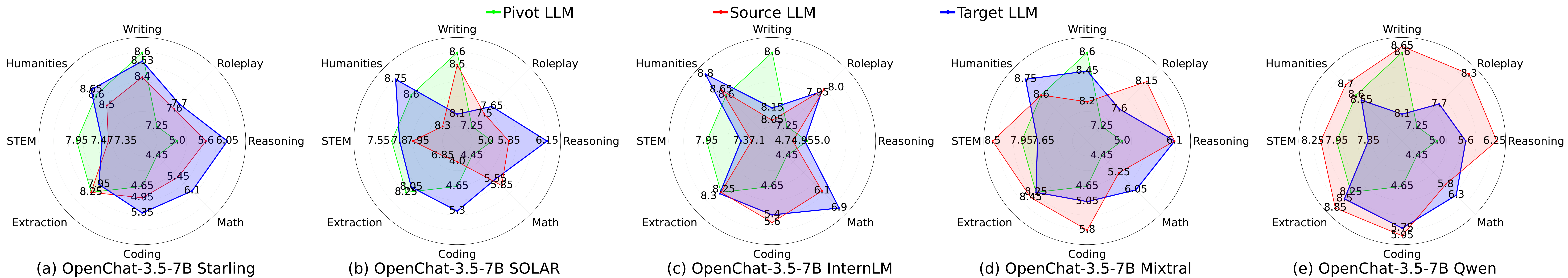}
	\caption{The effect of pairwise knowledge fusion for source LLMs across various domains on MT-Bench. It combines the strengths of each source LLM and the pivot (\texttt{OpenChat-3.5-7B}) into a more potent target LLM.}
	\label{fig:mt_bench_radar_multi}
	\vspace{-0.6cm}
\end{figure*}

\begin{wrapfigure}{r}{0.38\textwidth}
    \vspace{-0.4cm}
    \centering
    \includegraphics[width=0.98\linewidth]{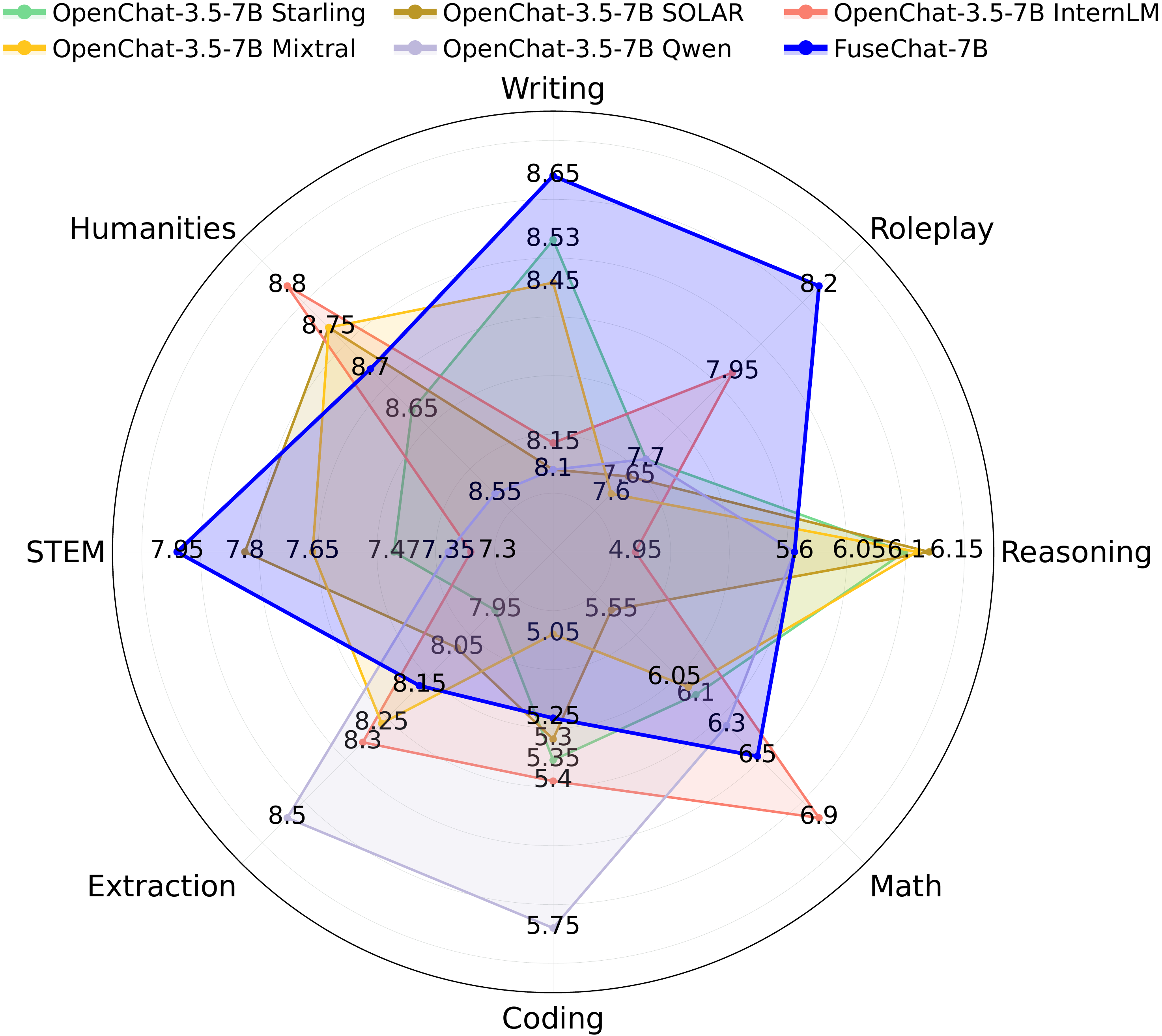}
    \caption{The effect of merging target LLMs into \texttt{\textsc{FuseChat-7B}} to combine their strengths across domains on MT-Bench.}
    \label{fig:mt_bench_radar_single}
    \vspace{-0.3cm}
\end{wrapfigure}

To further illustrate that our performance improvements stem from the integration of distinct knowledge from multiple LLMs, we evaluate the source LLMs, target LLMs, and \textsc{FuseChat} across various domains on MT-Bench. 
The results in Figure \ref{fig:mt_bench_radar_multi} reveal that the target LLMs demonstrate noticeable performance enhancements in most domains after pairwise knowledge fusion.
Typically, the performance of each target LLM falls between that of the pivot LLM and the respective source LLM.
This phenomenon can be attributed to the fusion function we employed to select the optimal target distributions with minimal cross-entropy, which promotes the incorporation of unique advantages from the pivot LLM and source LLMs into more potent target LLMs.
Notably, in math and reasoning domains, the performance of certain target LLMs surpasses that of either the pivot or source LLMs. This observation can be explained by the training data used for knowledge fusion, which enhances the target LLMs through a process akin to continual fine-tuning. It is also consistent with findings from knowledge distillation~\citep{wu2023ad}, where the student model occasionally outperforms the teacher in specific tasks. 
The effect of further merging these target LLMs into \texttt{\textsc{FuseChat-7B}} is shown in Figure \ref{fig:mt_bench_radar_single}. By integrating the capabilities of the target LLMs, \textsc{FuseChat} achieves a balanced and robust performance across diverse domains.

\subsection{Different Pivot LLM}
\vspace{-0.1cm}
\begin{wrapfigure}{r}{0.4\textwidth}
\centering
\vspace{-0.4cm}
\resizebox{0.99\linewidth}{!}{
    \begin{tabular}{lcc}
    \toprule
        \textbf{Model} & \textbf{AlpacaEval 2.0} & \textbf{MT-Bench} \\
        \midrule
        Starling-LM-7B-alpha & 14.70 & 7.01 \\
        Starling-LM-7B-alpha SFT & 13.20  (\textcolor{darkred}{-10.20\%}) & 6.89  (\textcolor{darkred}{-1.71\%}) \\ \midrule
        \textsc{FuseChat}-Starling-7B & 17.29  (\textcolor{darkblue}{+17.62\%}) & 7.16  (\textcolor{darkblue}{+2.14\%}) \\
        \bottomrule
    \end{tabular}
}
\captionsetup{type=table}
\caption{Starling-LM-7B-alpha as pivot LLM results on AlpacaEval 2.0 and MT-Bench.}
\label{tab:starling_pivot}
\vspace{-0.3cm}
\end{wrapfigure}

We conduct experiments using \texttt{Starling-LM-7B-alpha} to replace \texttt{OpenChat-3.5-7B} as a more robust pivot LLM, which achieved an LC Win Rate of 14.70 on AlpacaEval 2.0 and an Average Score of 7.01 on MT-Bench. The evaluation results presented in Table \ref{tab:starling_pivot} show that \texttt{\textsc{FuseChat}-Starling-7B} outperforms \texttt{Starling-LM-7B-alpha}, with relative performance improvements of 17.62\% on AlpacaEval 2.0 and 2.14\% on MT-Bench. Notably, although \texttt{Starling-} \texttt{LM-7B-alpha SFT} does not result in performance gains, the pairwise knowledge fusion and model merging processes lead to significant enhancements using the same training data.

\vspace{-0.15cm}
\subsection{Dataset Scaling}
\vspace{-0.1cm}
\begin{wrapfigure}{r}{0.4\textwidth}
\centering
\vspace{-0.4cm}
\resizebox{0.999\linewidth}{!}{
    \begin{tabular}{llccccc}
    \toprule
        \multirow{3}{*}{\textbf{Dataset Scale}}  &\multirow{3}{*} & \multicolumn{3}{c}{\textbf{MT-Bench}} \\   \cmidrule(lr){3-5}
        ~ & ~ &  \textbf{1st Turn} & \textbf{2nd Turn} & \textbf{Average Score} \\
        \midrule
        10,000  & ~ & 7.34 & 6.86 & 7.10 \\ 
        25,000 & ~  & 7.58 & 6.85 & 7.21 \\ 
        95,000  & ~ & \textbf{7.70} & \textbf{7.05} & \textbf{7.38} \\ 
        \bottomrule 
    \end{tabular}
}
\captionsetup{type=table}
\caption{Comparison of different dataset scales on MT-Bench.}
\label{tab:dataset_scale}
\vspace{-0.2cm}
\end{wrapfigure}

We perform experiments across different dataset scales for pairwise knowledge fusion, followed by merging the resulting target LLMs in the parameter space to obtain the final fused LLM. The results in Table \ref{tab:dataset_scale} indicate that the performance of the final fused LLM consistently improves as the training data scales up from 10k to 95k on MT-Bench, demonstrating the potential effectiveness of scaling up dataset to our method. 
\vspace{-0.15cm}
\subsection{Analysis of Model Merging}
\vspace{-0.1cm}
\label{sec:analysis_merge}
\begin{wrapfigure}{r}{0.4\textwidth}
\centering
\vspace{-0.4cm}
\resizebox{0.999\linewidth}{!}{
    \begin{tabular}{lcc}
    \toprule
        \textbf{Model} & \textbf{AlpacaEval 2.0} & \textbf{MT-Bench} \\
        \midrule
        \textsc{FuseChat-7B} Linear & \underline{17.12} & 7.03 \\
        \textsc{FuseChat-7B} TA & 15.74 & 7.08  \\
        \textsc{FuseChat-7B} TIES & 16.55 & \underline{7.33}  \\
        \textsc{FuseChat-7B} DARE & 16.57 & 7.15 \\ \midrule
        \textsc{FuseChat-7B} SCE & \textbf{17.16} & \textbf{7.38} \\
        \bottomrule
    \end{tabular}
}
\captionsetup{type=table}
\caption{Comparison of different merging methods on AlpacaEval 2.0 and MT-Bench.}
\label{tab:merge_res}
\vspace{-0.2cm}
\end{wrapfigure}

To investigate the effectiveness of the proposed SCE approach, we incorporate the target LLMs using different merging methods, including Linear~\citep{wortsman2022model}, TA~\citep{ilharco2023editing}, TIES~\citep{yadav2024ties}, and DARE~\citep{yu2023language}.
We evaluate the performance of these merged LLMs on AlpacaEval 2.0 and MT-Bench.
As depicted in Table \ref{tab:merge_res}, \texttt{\textsc{FuseChat-7B} SCE} outperforms all baseline merging methods on the two benchmarks.
For more details of baselines and hyperparameters in model merging, please refer to Appendix \ref{appendix: details_of_model_merging}.

\begin{wrapfigure}{r}{0.4\textwidth}
    \vspace{-0.4cm}
    \centering
    \includegraphics[width=0.999\linewidth]{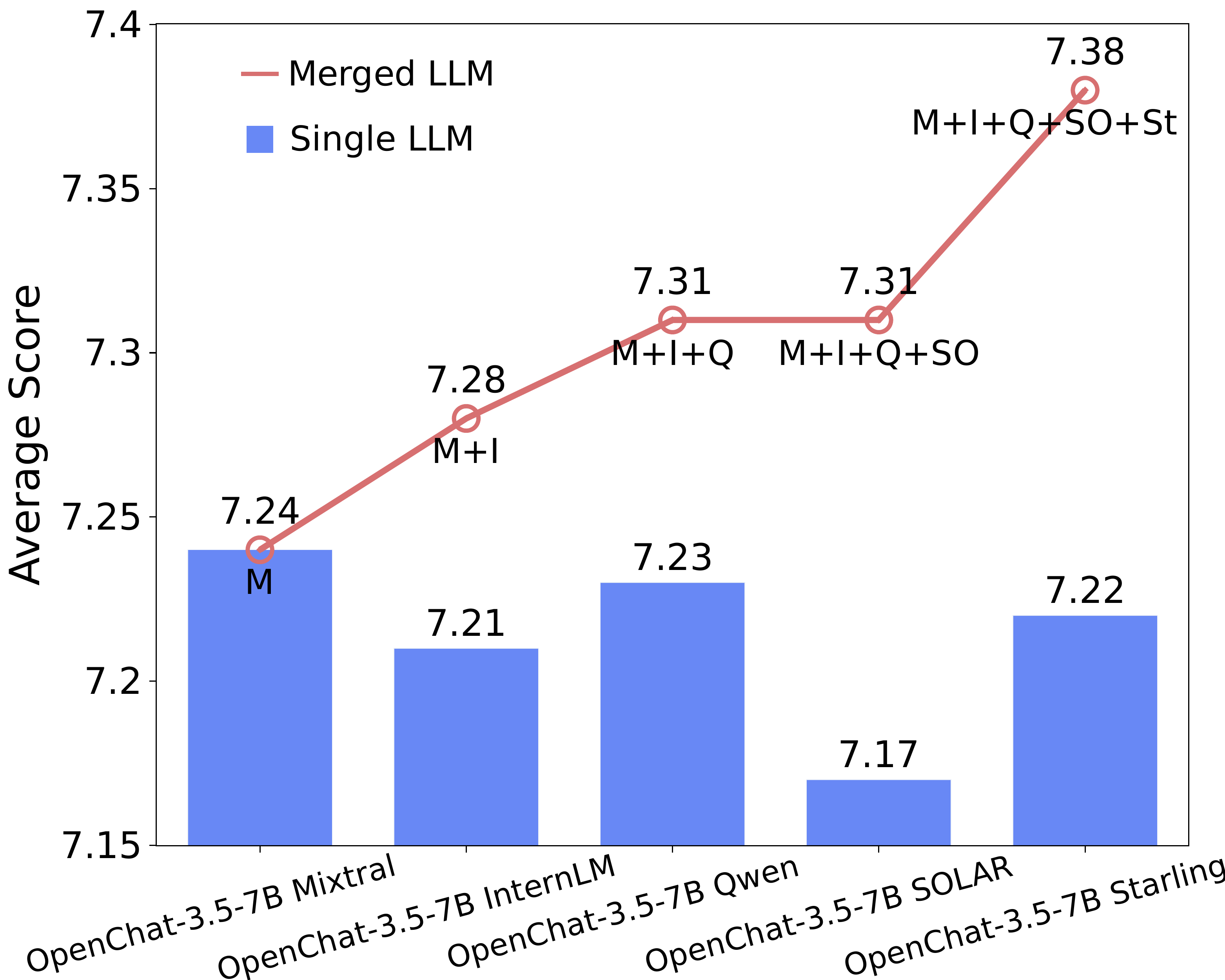}
    \caption{Results of \textsc{FuseChat} by merging varying numbers of target LLMs.}
    \label{fig:number}
    \vspace{-0.3cm}
\end{wrapfigure}

In Figure \ref{fig:number}, we further illustrate the performance of \texttt{\textsc{FuseChat-7B} SCE} by incorporating varying numbers of target LLMs on MT-Bench. The findings demonstrate a progressive enhancement in Average Score, which increases from 7.24 to 7.38 as the number of integrated target LLMs rises from 1 to 5. Moreover, we observe that after the integration of \texttt{OpenChat-3.5-7B SOLAR}, the performance of the merged LLM remains stable. This stabilization might be attributed to the comparatively sub-optimal performance of \texttt{OpenChat-3.5-7B SOLAR} and its corresponding \texttt{NH2-SOLAR-10.7B} compared to other target or source LLMs. Therefore, we suggest that both the diversity and quality of integrated source LLMs are critical factors for optimal knowledge fusion.
\vspace{-0.15cm}
\subsection{Ablation Studies for SCE}
\vspace{-0.1cm}
\begin{wrapfigure}{r}{0.4\textwidth}
\centering
\vspace{-0.4cm}
\resizebox{0.99\linewidth}{!}{
    \begin{tabular}{lcc}
    \toprule
        \textbf{Model} & \textbf{AlpacaEval 2.0} & \textbf{MT-Bench} \\
        \midrule
        \textsc{FuseChat-7B} SCE & 17.16 & 7.38 \\
        \textsc{FuseChat-7B} CE & 15.69  (\textcolor{darkred}{-8.57\%}) & 7.29  (\textcolor{darkred}{-1.22\%}) \\
        \textsc{FuseChat-7B} C & 16.62  (\textcolor{darkred}{-3.15\%}) & 7.11  (\textcolor{darkred}{-3.66\%}) \\
        \bottomrule
    \end{tabular}
}
\captionsetup{type=table}
\caption{Comparison of different merging methods on AlpacaEval 2.0 and MT-Bench. ``CE'' and ``C'' mean only the \emph{calculate}\&\emph{erase} and \emph{calculate} operations are used.}
\label{tab:ablation_res}
\vspace{-0.5cm}
\end{wrapfigure}

In this section, we conduct experiments to examine the effectiveness of the \emph{select}, \emph{calculate}, and \emph{erase} operations in SCE. 
The results in Table \ref{tab:ablation_res} demonstrate that, without the \emph{select} operations, \texttt{\textsc{FuseChat-7B} CE} suffers substantial performance degradation on the two benchmarks. This underscores the benefits of selecting salient elements from fusion vectors with high variance among target LLMs to signify their distinctive and significant strengths. Moreover, removing both the \emph{select} and \emph{erase} operations leads to \texttt{\textsc{FuseChat-7B} C} with decreased performance on MT-Bench, highlighting the importance of resolving parameter interference in fusion vectors from different target LLMs.

\subsection{Analysis of Token Alignment}
\begin{wrapfigure}{r}{0.40\textwidth}
    \vspace{-0.8cm}
    \centering
    \includegraphics[width=0.98\linewidth]{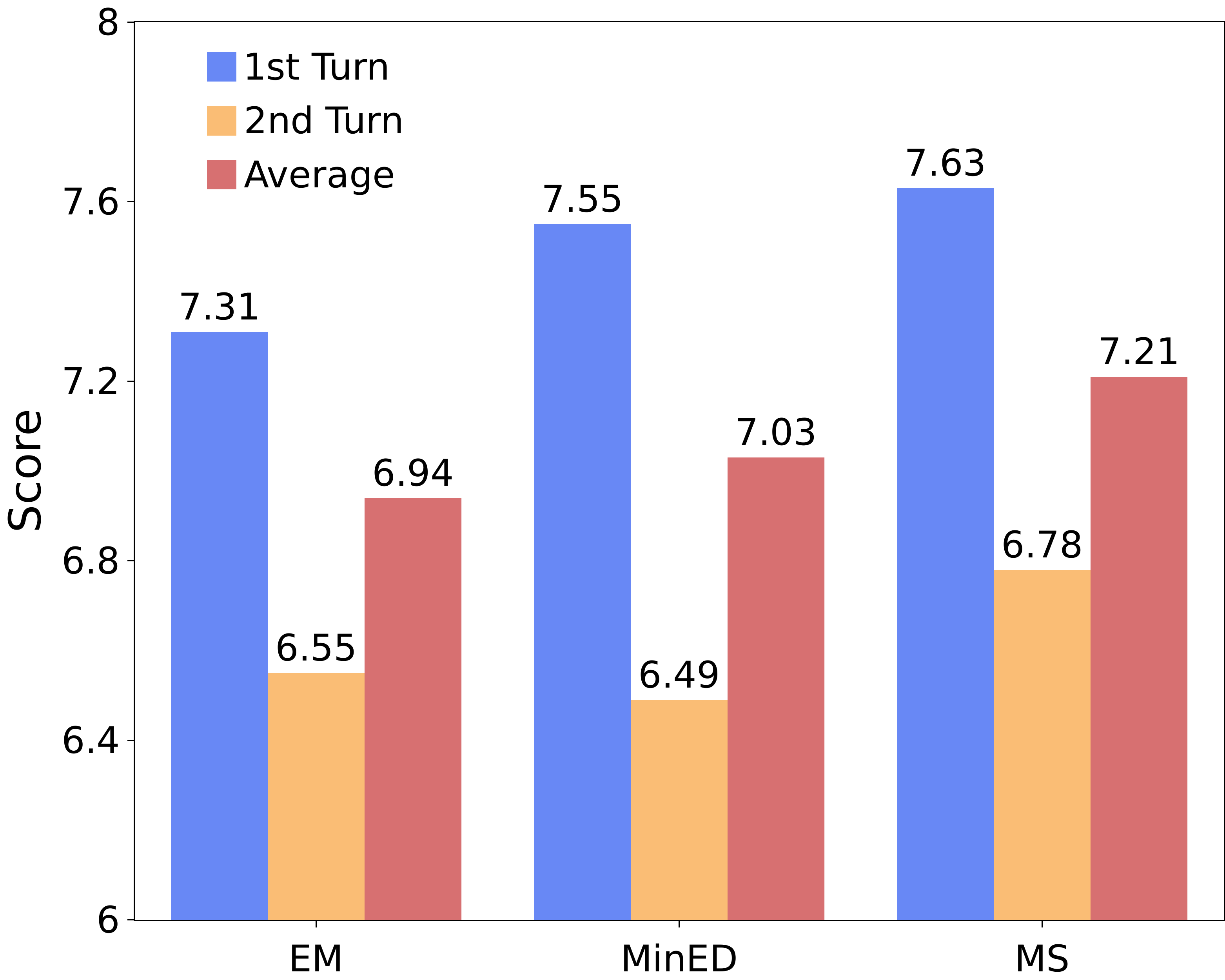}
    \caption{Results of \texttt{OpenChat-3.5-7B InternLM} via pairwise knowledge fusion with different token alignment strategies.}
    \label{fig:token}
    \vspace{-0.3cm}
\end{wrapfigure}

Finally, we delve into exploring the impact of various token alignment strategies. Specifically, we apply EM~\citep{fu2023specializing} and MinED~\citep{wan2024knowledge}, and our MS methods to align distributions generated by \texttt{InterLM2-Chat-20B} with those of \texttt{OpenChat-3.5-7B}. Then, we conduct pairwise knowledge fusion to derive \texttt{OpenChat-3.5-7B InternLM}. As depicted in Figure \ref{fig:token}, our proposed MS method, rooted in mapping statistics, consistently outperforms EM and MinED, which rely on exact matching and minimal edit distance, respectively. We propose that this performance enhancement arises from MS's effective utilization of token mapping statistics within the dataset, which greatly improves the effect of token alignment in the distribution dimension.

\section{Conclusion}
\label{sec:conclusion}
In this work, we propose a fuse-and-merge framework for knowledge fusion of structurally and scale-varied chat LLMs to integrate their collective knowledge and individual strengths into a more potent chat LLM, resulting in \textsc{FuseChat}. 
\textsc{FuseChat} first undertakes pairwise knowledge fusion for source chat LLMs to derive multiple target LLMs of identical structure and size via lightweight fine-tuning. 
Then, these target LLMs are merged within the parameter space using a novel method SCE to calculate the merging coefficients based on the magnitude of parameter updates before and after fine-tuning.
Experimental results on two representative instruction-following benchmarks demonstrate the superiority of \textsc{FuseChat} across different model scales, even comparable to \texttt{Mixtral-8x7B-Instruct} and approaching \texttt{GPT-3.5-Turbo-1106} on MT-Bench.

The concept of knowledge fusion shares similarities with related approaches, such as the recently popular mixture of experts (MoEs). Both methods aim to leverage the strengths of multiple models (experts). However, while MoEs require loading multiple experts during inference, leading to higher time and memory requirements, knowledge fusion allows the integration of multiple LLMs with diverse architectures and scales into a single LLM without additional time or memory overhead. This makes knowledge fusion more efficient, especially when model size is a critical consideration.

\section{Limitations and Future Work}
\label{sec:limitations}
Our work relies on constructing a knowledge fusion dataset that spans diverse domains and leverages the strengths of source LLMs. This process demands substantial data engineering efforts, which may limit the scalability of our methodology. Future research should focus on developing more efficient data synthesis techniques to expand the scope of the knowledge fusion dataset. Additionally, while our study shows improvements in chat model capabilities, it does not address other critical aspects of LLMs, such as knowledge comprehension and the mitigation of hallucinations. Further investigation is necessary to evaluate the applicability and effectiveness of our approach in these areas.

\bibliographystyle{apalike}
\bibliography{custom}

\begin{thebibliography}{}

\bibitem[Achiam et~al., 2023]{achiam2023gpt}
Achiam, J., Adler, S., Agarwal, S., Ahmad, L., Akkaya, I., Aleman, F.~L., Almeida, D., Altenschmidt, J., Altman, S., Anadkat, S., et~al. (2023).
\newblock Gpt-4 technical report.
\newblock {\em arXiv preprint arXiv:2303.08774}.

\bibitem[Agarwal et~al., 2024]{agarwal2024onpolicy}
Agarwal, R., Vieillard, N., Zhou, Y., Stanczyk, P., Garea, S.~R., Geist, M., and Bachem, O. (2024).
\newblock On-policy distillation of language models: Learning from self-generated mistakes.
\newblock In {\em The Twelfth International Conference on Learning Representations}.

\bibitem[Akiba et~al., 2024]{akiba2024evolutionary}
Akiba, T., Shing, M., Tang, Y., Sun, Q., and Ha, D. (2024).
\newblock Evolutionary optimization of model merging recipes.
\newblock {\em arXiv preprint arXiv:2403.13187}.

\bibitem[Anthropic, 2024]{anthropic2024claude}
Anthropic, A. (2024).
\newblock The claude 3 model family: Opus, sonnet, haiku.
\newblock {\em Claude-3 Model Card}.

\bibitem[Bai et~al., 2023]{bai2023qwen}
Bai, J., Bai, S., Chu, Y., Cui, Z., Dang, K., Deng, X., Fan, Y., Ge, W., Han, Y., Huang, F., et~al. (2023).
\newblock Qwen technical report.
\newblock {\em arXiv preprint arXiv:2309.16609}.

\bibitem[Bisk et~al., 2020]{bisk2020piqa}
Bisk, Y., Zellers, R., Gao, J., Choi, Y., et~al. (2020).
\newblock Piqa: Reasoning about physical commonsense in natural language.
\newblock In {\em Proceedings of the AAAI Conference on Artificial Intelligence}, pages 7432--7439.

\bibitem[Brown et~al., 2020]{brown2020language}
Brown, T., Mann, B., Ryder, N., Subbiah, M., Kaplan, J.~D., Dhariwal, P., Neelakantan, A., Shyam, P., Sastry, G., Askell, A., et~al. (2020).
\newblock Language models are few-shot learners.
\newblock {\em Advances in neural information processing systems}, 33:1877--1901.

\bibitem[Cai et~al., 2024]{cai2024internlm2}
Cai, Z., Cao, M., Chen, H., Chen, K., Chen, K., Chen, X., Chen, X., Chen, Z., Chen, Z., Chu, P., et~al. (2024).
\newblock Internlm2 technical report.
\newblock {\em arXiv preprint arXiv:2403.17297}.

\bibitem[Clark et~al., 2019]{clark2019boolq}
Clark, C., Lee, K., Chang, M.-W., Kwiatkowski, T., Collins, M., and Toutanova, K. (2019).
\newblock Boolq: Exploring the surprising difficulty of natural yes/no questions.
\newblock In {\em Proceedings of the 2019 Conference of the North American Chapter of the Association for Computational Linguistics: Human Language Technologies, Volume 1 (Long and Short Papers)}, pages 2924--2936.

\bibitem[Cobbe et~al., 2021]{cobbe2021training}
Cobbe, K., Kosaraju, V., Bavarian, M., Chen, M., Jun, H., Kaiser, L., Plappert, M., Tworek, J., Hilton, J., Nakano, R., et~al. (2021).
\newblock Training verifiers to solve math word problems.
\newblock {\em arXiv preprint arXiv:2110.14168}.

\bibitem[Ding et~al., 2024]{ding2024mastering}
Ding, N., Chen, Y., Cui, G., Lv, X., Xie, R., Zhou, B., Liu, Z., and Sun, M. (2024).
\newblock Mastering text, code and math simultaneously via fusing highly specialized language models.
\newblock {\em arXiv preprint arXiv:2403.08281}.

\bibitem[Dubois et~al., 2024a]{dubois2024length}
Dubois, Y., Galambosi, B., Liang, P., and Hashimoto, T.~B. (2024a).
\newblock Length-controlled alpacaeval: A simple way to debias automatic evaluators.
\newblock {\em arXiv preprint arXiv:2404.04475}.

\bibitem[Dubois et~al., 2024b]{dubois2024alpacafarm}
Dubois, Y., Li, C.~X., Taori, R., Zhang, T., Gulrajani, I., Ba, J., Guestrin, C., Liang, P.~S., and Hashimoto, T.~B. (2024b).
\newblock Alpacafarm: A simulation framework for methods that learn from human feedback.
\newblock {\em Advances in Neural Information Processing Systems}, 36.

\bibitem[Fedus et~al., 2022]{fedus2022switch}
Fedus, W., Zoph, B., and Shazeer, N. (2022).
\newblock Switch transformers: Scaling to trillion parameter models with simple and efficient sparsity.
\newblock {\em Journal of Machine Learning Research}, 23(120):1--39.

\bibitem[Feng et~al., 2024]{feng2024mixture}
Feng, W., Hao, C., Zhang, Y., Han, Y., and Wang, H. (2024).
\newblock Mixture-of-loras: An efficient multitask tuning for large language models.
\newblock {\em arXiv preprint arXiv:2403.03432}.

\bibitem[Fisher, 1922]{fisher1922mathematical}
Fisher, R.~A. (1922).
\newblock On the mathematical foundations of theoretical statistics.
\newblock {\em Philosophical transactions of the Royal Society of London. Series A, containing papers of a mathematical or physical character}, 222(594-604):309--368.

\bibitem[Fu et~al., 2023]{fu2023specializing}
Fu, Y., Peng, H., Ou, L., Sabharwal, A., and Khot, T. (2023).
\newblock Specializing smaller language models towards multi-step reasoning.
\newblock In {\em International Conference on Machine Learning}, pages 10421--10430. PMLR.

\bibitem[Gu et~al., 2024]{gu2024minillm}
Gu, Y., Dong, L., Wei, F., and Huang, M. (2024).
\newblock Mini{LLM}: Knowledge distillation of large language models.
\newblock In {\em The Twelfth International Conference on Learning Representations}.

\bibitem[Gupta et~al., 2020]{gupta2020stochastic}
Gupta, V., Serrano, S.~A., and DeCoste, D. (2020).
\newblock Stochastic weight averaging in parallel: Large-batch training that generalizes well.
\newblock {\em International Conference on Learning Representations}.

\bibitem[Hendrycks et~al., 2021]{hendrycks2021measuring}
Hendrycks, D., Burns, C., Kadavath, S., Arora, A., Basart, S., Tang, E., Song, D., and Steinhardt, J. (2021).
\newblock Measuring mathematical problem solving with the math dataset.
\newblock In {\em Thirty-fifth Conference on Neural Information Processing Systems Datasets and Benchmarks Track (Round 2)}.

\bibitem[Hinton et~al., 2015]{hinton2015distilling}
Hinton, G., Vinyals, O., and Dean, J. (2015).
\newblock Distilling the knowledge in a neural network.
\newblock {\em arXiv preprint arXiv:1503.02531}.

\bibitem[Hu et~al., 2022]{hu2022lora}
Hu, E.~J., Wallis, P., Allen-Zhu, Z., Li, Y., Wang, S., Wang, L., Chen, W., et~al. (2022).
\newblock Lora: Low-rank adaptation of large language models.
\newblock In {\em International Conference on Learning Representations}.

\bibitem[Ilharco et~al., 2023]{ilharco2023editing}
Ilharco, G., Ribeiro, M.~T., Wortsman, M., Schmidt, L., Hajishirzi, H., and Farhadi, A. (2023).
\newblock Editing models with task arithmetic.
\newblock In {\em The Eleventh International Conference on Learning Representations}.

\bibitem[Jiang et~al., 2024]{jiang2024mixtral}
Jiang, A.~Q., Sablayrolles, A., Roux, A., Mensch, A., Savary, B., Bamford, C., Chaplot, D.~S., Casas, D. d.~l., Hanna, E.~B., Bressand, F., et~al. (2024).
\newblock Mixtral of experts.
\newblock {\em arXiv preprint arXiv:2401.04088}.

\bibitem[Jiang et~al., 2023]{jiang2023llm}
Jiang, D., Ren, X., and Lin, B.~Y. (2023).
\newblock Llm-blender: Ensembling large language models with pairwise ranking and generative fusion.
\newblock In {\em Proceedings of the 61st Annual Meeting of the Association for Computational Linguistics (Volume 1: Long Papers)}, pages 14165--14178.

\bibitem[Jiao et~al., 2020]{jiao2020tinybert}
Jiao, X., Yin, Y., Shang, L., Jiang, X., Chen, X., Li, L., Wang, F., and Liu, Q. (2020).
\newblock Tinybert: Distilling bert for natural language understanding.
\newblock In {\em Findings of the Association for Computational Linguistics: EMNLP 2020}, pages 4163--4174.

\bibitem[Jin et~al., 2023]{jin2023dataless}
Jin, X., Ren, X., Preotiuc-Pietro, D., and Cheng, P. (2023).
\newblock Dataless knowledge fusion by merging weights of language models.
\newblock In {\em The Eleventh International Conference on Learning Representations}.

\bibitem[Khanuja et~al., 2021]{khanuja2021mergedistill}
Khanuja, S., Johnson, M., and Talukdar, P. (2021).
\newblock Mergedistill: Merging language models using pre-trained distillation.
\newblock In {\em Findings of the Association for Computational Linguistics: ACL-IJCNLP 2021}, pages 2874--2887.

\bibitem[Kim et~al., 2023]{kim2023solar}
Kim, D., Park, C., Kim, S., Lee, W., Song, W., Kim, Y., Kim, H., Kim, Y., Lee, H., Kim, J., et~al. (2023).
\newblock Solar 10.7 b: Scaling large language models with simple yet effective depth up-scaling.
\newblock {\em arXiv preprint arXiv:2312.15166}.

\bibitem[Kim and Rush, 2016]{kim2016sequence}
Kim, Y. and Rush, A.~M. (2016).
\newblock Sequence-level knowledge distillation.
\newblock In {\em Proceedings of the 2016 Conference on Empirical Methods in Natural Language Processing}, pages 1317--1327.

\bibitem[Komatsuzaki et~al., 2023]{komatsuzaki2023sparse}
Komatsuzaki, A., Puigcerver, J., Lee-Thorp, J., Ruiz, C.~R., Mustafa, B., Ainslie, J., Tay, Y., Dehghani, M., and Houlsby, N. (2023).
\newblock Sparse upcycling: Training mixture-of-experts from dense checkpoints.
\newblock In {\em The Eleventh International Conference on Learning Representations}.

\bibitem[K{\"o}pf et~al., 2024]{kopf2024openassistant}
K{\"o}pf, A., Kilcher, Y., von R{\"u}tte, D., Anagnostidis, S., Tam, Z.~R., Stevens, K., Barhoum, A., Nguyen, D., Stanley, O., Nagyfi, R., et~al. (2024).
\newblock Openassistant conversations-democratizing large language model alignment.
\newblock {\em Advances in Neural Information Processing Systems}, 36.

\bibitem[Littlestone and Warmuth, 1994]{littlestone1994weighted}
Littlestone, N. and Warmuth, M.~K. (1994).
\newblock The weighted majority algorithm.
\newblock {\em Information and Computation}, 108(2):212--261.

\bibitem[Loshchilov and Hutter, 2019]{loshchilov2019decoupled}
Loshchilov, I. and Hutter, F. (2019).
\newblock Decoupled weight decay regularization.
\newblock In {\em International Conference on Learning Representations}.

\bibitem[Luo et~al., 2024]{luo2024wizardcoder}
Luo, Z., Xu, C., Zhao, P., Sun, Q., Geng, X., Hu, W., Tao, C., Ma, J., Lin, Q., and Jiang, D. (2024).
\newblock Wizardcoder: Empowering code large language models with evol-instruct.
\newblock In {\em The Twelfth International Conference on Learning Representations}.

\bibitem[Matena and Raffel, 2022]{matena2022merging}
Matena, M.~S. and Raffel, C.~A. (2022).
\newblock Merging models with fisher-weighted averaging.
\newblock {\em Advances in Neural Information Processing Systems}, 35:17703--17716.

\bibitem[Mavromatis et~al., 2024]{mavromatis2024pack}
Mavromatis, C., Karypis, P., and Karypis, G. (2024).
\newblock Pack of llms: Model fusion at test-time via perplexity optimization.
\newblock {\em arXiv preprint arXiv:2404.11531}.

\bibitem[Meng et~al., 2024]{meng2024simpo}
Meng, Y., Xia, M., and Chen, D. (2024).
\newblock Simpo: Simple preference optimization with a reference-free reward.
\newblock {\em arXiv preprint arXiv:2405.14734}.

\bibitem[Monteith et~al., 2011]{monteith2011turning}
Monteith, K., Carroll, J.~L., Seppi, K., and Martinez, T. (2011).
\newblock Turning bayesian model averaging into bayesian model combination.
\newblock In {\em The 2011 International Joint Conference on Neural Networks}, pages 2657--2663. IEEE.

\bibitem[Mukherjee et~al., 2023]{mukherjee2023orca}
Mukherjee, S., Mitra, A., Jawahar, G., Agarwal, S., Palangi, H., and Awadallah, A. (2023).
\newblock Orca: Progressive learning from complex explanation traces of gpt-4.
\newblock {\em arXiv preprint arXiv:2306.02707}.

\bibitem[Peng et~al., 2023]{peng2023instruction}
Peng, B., Li, C., He, P., Galley, M., and Gao, J. (2023).
\newblock Instruction tuning with gpt-4.
\newblock {\em arXiv preprint arXiv:2304.03277}.

\bibitem[Rein et~al., 2023]{rein2023gpqa}
Rein, D., Hou, B.~L., Stickland, A.~C., Petty, J., Pang, R.~Y., Dirani, J., Michael, J., and Bowman, S.~R. (2023).
\newblock Gpqa: A graduate-level google-proof q\&a benchmark.
\newblock {\em arXiv preprint arXiv:2311.12022}.

\bibitem[Sanh et~al., 2019]{sanh2019distilbert}
Sanh, V., Debut, L., Chaumond, J., and Wolf, T. (2019).
\newblock Distilbert, a distilled version of bert: smaller, faster, cheaper and lighter.
\newblock {\em arXiv preprint arXiv:1910.01108}.

\bibitem[Sukhbaatar et~al., 2024]{sukhbaatar2024branch}
Sukhbaatar, S., Golovneva, O., Sharma, V., Xu, H., Lin, X.~V., Rozi{\`e}re, B., Kahn, J., Li, D., Yih, W.-t., Weston, J., et~al. (2024).
\newblock Branch-train-mix: Mixing expert llms into a mixture-of-experts llm.
\newblock {\em arXiv preprint arXiv:2403.07816}.

\bibitem[Sun et~al., 2019]{sun2019patient}
Sun, S., Cheng, Y., Gan, Z., and Liu, J. (2019).
\newblock Patient knowledge distillation for bert model compression.
\newblock In {\em Proceedings of the 2019 Conference on Empirical Methods in Natural Language Processing and the 9th International Joint Conference on Natural Language Processing (EMNLP-IJCNLP)}, pages 4323--4332.

\bibitem[Touvron et~al., 2023]{touvron2023llama-2}
Touvron, H., Martin, L., Stone, K., Albert, P., Almahairi, A., Babaei, Y., Bashlykov, N., Batra, S., Bhargava, P., Bhosale, S., et~al. (2023).
\newblock Llama 2: Open foundation and fine-tuned chat models.
\newblock {\em arXiv preprint arXiv:2307.09288}.

\bibitem[Turc et~al., 2019]{turc2019well}
Turc, I., Chang, M.-W., Lee, K., and Toutanova, K. (2019).
\newblock Well-read students learn better: On the importance of pre-training compact models.
\newblock {\em arXiv preprint arXiv:1908.08962}.

\bibitem[Wan et~al., 2024]{wan2024knowledge}
Wan, F., Huang, X., Cai, D., Quan, X., Bi, W., and Shi, S. (2024).
\newblock Knowledge fusion of large language models.
\newblock In {\em The Twelfth International Conference on Learning Representations}.

\bibitem[Wang et~al., 2024a]{wang2024openchat}
Wang, G., Cheng, S., Zhan, X., Li, X., Song, S., and Liu, Y. (2024a).
\newblock Openchat: Advancing open-source language models with mixed-quality data.
\newblock In {\em The Twelfth International Conference on Learning Representations}.

\bibitem[Wang et~al., 2020]{wang2020minilm}
Wang, W., Wei, F., Dong, L., Bao, H., Yang, N., and Zhou, M. (2020).
\newblock Minilm: Deep self-attention distillation for task-agnostic compression of pre-trained transformers.
\newblock {\em Advances in Neural Information Processing Systems}, 33:5776--5788.

\bibitem[Wang et~al., 2024b]{wang2024mmlu}
Wang, Y., Ma, X., Zhang, G., Ni, Y., Chandra, A., Guo, S., Ren, W., Arulraj, A., He, X., Jiang, Z., et~al. (2024b).
\newblock Mmlu-pro: A more robust and challenging multi-task language understanding benchmark.
\newblock {\em arXiv preprint arXiv:2406.01574}.

\bibitem[Wei et~al., 2023]{wei2023magicoder}
Wei, Y., Wang, Z., Liu, J., Ding, Y., and Zhang, L. (2023).
\newblock Magicoder: Source code is all you need.
\newblock {\em arXiv preprint arXiv:2312.02120}.

\bibitem[Wolf et~al., 2020]{wolf2020transformers}
Wolf, T., Debut, L., Sanh, V., Chaumond, J., Delangue, C., Moi, A., Cistac, P., Rault, T., Louf, R., Funtowicz, M., et~al. (2020).
\newblock Transformers: State-of-the-art natural language processing.
\newblock In {\em Proceedings of the 2020 Conference on Empirical Methods in Natural Language Processing: System Dyiemonstrations}, pages 38--45.

\bibitem[Wortsman et~al., 2022]{wortsman2022model}
Wortsman, M., Ilharco, G., Gadre, S.~Y., Roelofs, R., Gontijo-Lopes, R., Morcos, A.~S., Namkoong, H., Farhadi, A., Carmon, Y., Kornblith, S., et~al. (2022).
\newblock Model soups: averaging weights of multiple fine-tuned models improves accuracy without increasing inference time.
\newblock In {\em International Conference on Machine Learning}, pages 23965--23998. PMLR.

\bibitem[Wu et~al., 2023]{wu2023ad}
Wu, S., Chen, H., Quan, X., Wang, Q., and Wang, R. (2023).
\newblock Ad-kd: Attribution-driven knowledge distillation for language model compression.
\newblock In {\em Proceedings of the 61st Annual Meeting of the Association for Computational Linguistics (Volume 1: Long Papers)}, pages 8449--8465.

\bibitem[Wu et~al., 2024]{wu2024self}
Wu, Y., Sun, Z., Yuan, H., Ji, K., Yang, Y., and Gu, Q. (2024).
\newblock Self-play preference optimization for language model alignment.
\newblock {\em arXiv preprint arXiv:2405.00675}.

\bibitem[Yadav et~al., 2024]{yadav2024ties}
Yadav, P., Tam, D., Choshen, L., Raffel, C.~A., and Bansal, M. (2024).
\newblock Ties-merging: Resolving interference when merging models.
\newblock {\em Advances in Neural Information Processing Systems}, 36.

\bibitem[You et~al., 2017]{you2017learning}
You, S., Xu, C., Xu, C., and Tao, D. (2017).
\newblock Learning from multiple teacher networks.
\newblock In {\em Proceedings of the 23rd ACM SIGKDD international conference on knowledge discovery and data mining}, pages 1285--1294.

\bibitem[Yu et~al., 2024]{yu2024metamath}
Yu, L., Jiang, W., Shi, H., YU, J., Liu, Z., Zhang, Y., Kwok, J., Li, Z., Weller, A., and Liu, W. (2024).
\newblock Metamath: Bootstrap your own mathematical questions for large language models.
\newblock In {\em The Twelfth International Conference on Learning Representations}.

\bibitem[Yu et~al., 2023]{yu2023language}
Yu, L., Yu, B., Yu, H., Huang, F., and Li, Y. (2023).
\newblock Language models are super mario: Absorbing abilities from homologous models as a free lunch.
\newblock {\em arXiv preprint arXiv:2311.03099}.

\bibitem[Zhang et~al., 2023]{zhang2023composing}
Zhang, J., Liu, J., He, J., et~al. (2023).
\newblock Composing parameter-efficient modules with arithmetic operation.
\newblock {\em Advances in Neural Information Processing Systems}, 36:12589--12610.

\bibitem[Zheng et~al., 2024]{zheng2024judging}
Zheng, L., Chiang, W.-L., Sheng, Y., Zhuang, S., Wu, Z., Zhuang, Y., Lin, Z., Li, Z., Li, D., Xing, E., et~al. (2024).
\newblock Judging llm-as-a-judge with mt-bench and chatbot arena.
\newblock {\em Advances in Neural Information Processing Systems}, 36.

\bibitem[Zhou et~al., 2023]{zhou2023instruction}
Zhou, J., Lu, T., Mishra, S., Brahma, S., Basu, S., Luan, Y., Zhou, D., and Hou, L. (2023).
\newblock Instruction-following evaluation for large language models.
\newblock {\em arXiv preprint arXiv:2311.07911}.

\bibitem[Zhu et~al., 2023]{zhu2023starling}
Zhu, B., Frick, E., Wu, T., Zhu, H., and Jiao, J. (2023).
\newblock Starling-7b: Improving llm helpfulness \& harmlessness with rlaif.

\end{thebibliography}

\appendix
\vspace{-0.15cm}
\section{Details of Token Alignment}
\vspace{-0.1cm}
\label{appendix: details_of_token_alignment}

\begin{figure*}[htbp]
    \centering
    \includegraphics[width=0.90\textwidth]{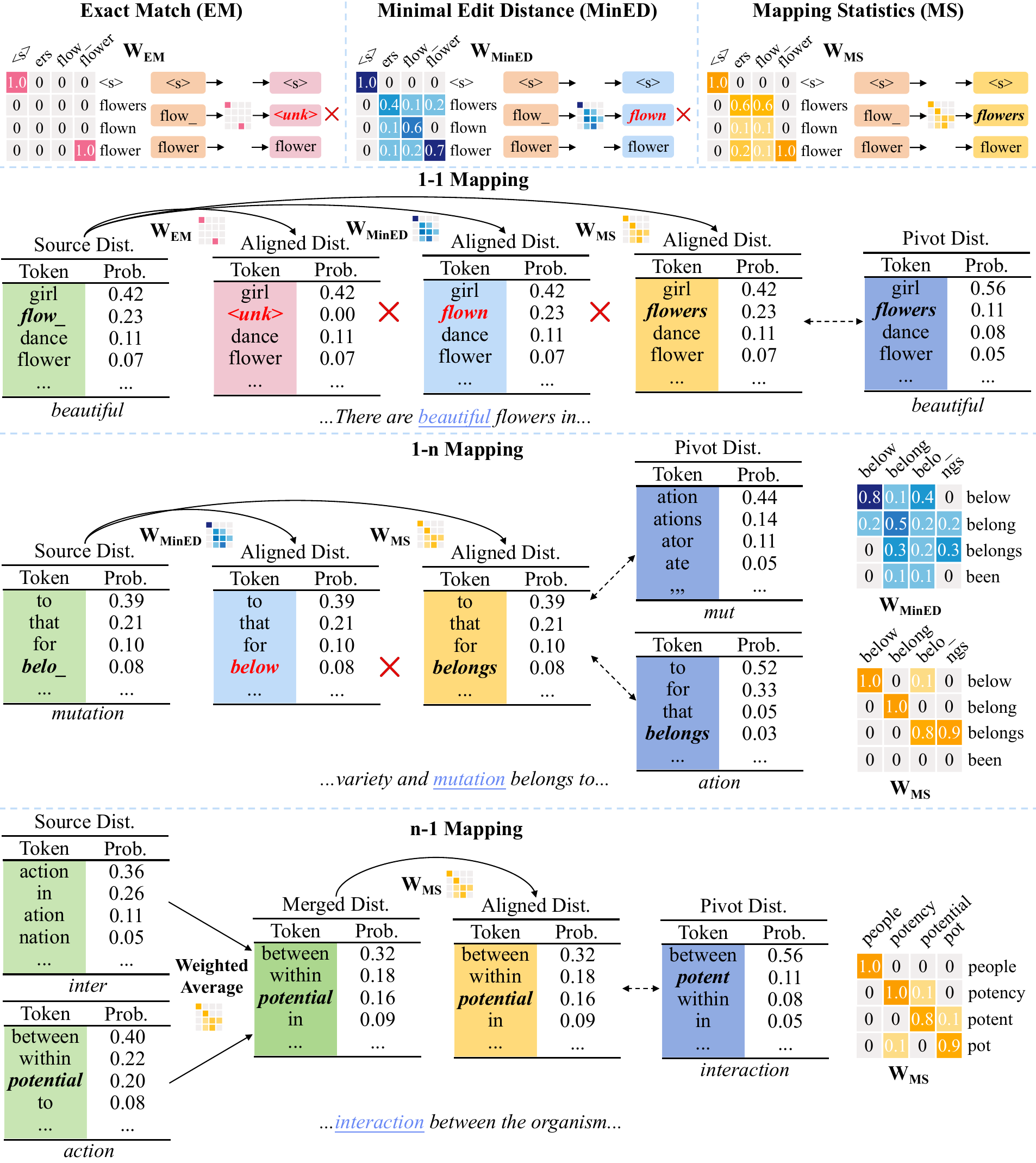}
     \caption{Illustration of EM, MinED, and our MS token alignment strategies in 1-1, 1-n, and n-1 mappings.}
	\label{fig:token_alignment_process}
	\vspace{-0.3cm}
\end{figure*}

In Figure \ref{fig:token_alignment_process}, we present the token pair mappings employed in three distinct token alignment strategies, including EM~\citep{fu2023specializing}, MinED~\citep{wan2024knowledge}, and our MS.
For clarity, these mapping strategies are depicted in a matrix format, where each column represents the probability of a source token being aligned with a corresponding pivot token.
The values within these matrices derive from the respective alignment strategies employed.
For instance, the matrix \(\mathbf{W_{EM}}\) relies on exact matches between source and pivot token pairs, while \(\mathbf{W_{MinED}}\) inversely relates to the edit distance between these pairs. \(\mathbf{W_{MS}}\) is based on the statistical mapping frequency between the source and pivot tokens.

In scenarios involving 1-1 or 1-n mappings, the EM and MinED methods utilize \(\mathbf{W_{EM}}\) or \(\mathbf{W_{MinED}}\) respectively, which may lead to inaccurate mappings.
For example, in EM, the token ``flow\_'' might be incorrectly aligned with ``<unk>'', and in MinED, ``flow\_'' could map to ``flown'', or ``belo\_'' to ``below''. 
In contrast, our MS method achieves more accurate alignments such as mapping ``flow\_'' to ``flowers'' and ``belo\_'' to ``belongs'', using \(\mathbf{W_{MS}}\) from sequence-dimensional token alignments.
For n-1 mapping, where only MS is applicable, multiple source distributions are combined using a weighted average determined by \(\mathbf{W_{MS}}\) to derive a merged distribution.
This unified distribution is then processed similarly to the 1-1 mappings.

\vspace{-0.15cm}
\section{Details of Model Merging}
\vspace{-0.1cm}
\label{appendix: details_of_model_merging}
The hyperparameters for various merging methods are detailed as follows. For the Linear method \citep{wortsman2022model}, merging parameters are calculated as the mean of all target LLMs. In the TA method \citep{ilharco2023editing}, we adhere to the original paper, exploring scaling coefficients within the range of {[}0.3, 0.4, 0.5{]}. The optimal setting of 0.3 is selected based on performance.
For the TIES \citep{yadav2024ties} and DARE \citep{yu2023language} approaches, we search for the trim/drop rate within the range of [0.1, 0.2, $\cdots$, 0.9]. The optimal trim/drop rate is 0.4, which results in the elimination of the bottom/random 40\% of delta parameters by resetting them to zero. Merging coefficients are computed as the average of all target LLMs.
For the SCE method, we search for the salient element selection thresholds $\tau$ within the range of [10, 20, $\cdots$, 90]. The optimal threshold is 10\%. Merging coefficients are automatically derived based on the magnitude of delta parameters.

\vspace{-0.15cm}
\section{Details of Training Dataset}
\vspace{-0.1cm}
\label{appendix: details_of_training_dataset}

We curated a comprehensive training dataset, \textsc{FuseChat-Mixture}, from various sources. 
This dataset covers different styles and capabilities, featuring both human-written and model-generated, and spanning general instruction-following and specific skills. These sources include: 

\textbf{Orca-Best}\footnote{\url{https://huggingface.co/datasets/shahules786/orca-best}}: We sampled 20,000 examples from Orca-Best, which is filtered from the GPT-4 (1M) partition of Orca~\citep{mukherjee2023orca} based on maximum length and clustering of instructions.

\textbf{Capybara}\footnote{\url{https://huggingface.co/datasets/LDJnr/Capybara}}: We incorporated all the 16,000 examples of Capybara, which is a high-quality collection of multi-turn synthetic conversations.

\textbf{No-Robots}\footnote{\url{https://huggingface.co/datasets/HuggingFaceH4/no_robots}}: We included all the 9,500 examples of No-Robots, which is a dataset created by skilled human annotators for supervised fine-tuning. 

\textbf{ShareGPT-GPT4}\footnote{\url{https://huggingface.co/datasets/shibing624/sharegpt_gpt4}}: We utilized all 6,200 examples from ShareGPT-GPT4, which exclusively uses dialogues generated by GPT-4 in ShareGPT.

\textbf{Oasst-Top1}\footnote{\url{https://huggingface.co/datasets/OpenAssistant/oasst_top1_2023-08-25}}: We selected 5,000 examples from Oasst-Top1, which is a refined version of Oasst1~\citep{kopf2024openassistant}, a human-annotated assistant-style conversation dataset.

\textbf{MetaMathQA}~\footnote{\url{https://huggingface.co/datasets/meta-math/MetaMathQA}}: We sampled 10,000 examples from MetaMathQA~\citep{yu2024metamath}, which is augmented from the GSM8K~\citep{cobbe2021training} and MATH~\citep{hendrycks2021measuring} datasets for mathematics problem-solving.

\textbf{OSS-Instruct}~\footnote{\url{https://huggingface.co/datasets/ise-uiuc/Magicoder-OSS-Instruct-75K}}: We chose 10,000 examples from OSS-Instruct~\citep{wei2023magicoder}, which contains code instruction data synthesized from open-source code snippets.

\textbf{Evol-Alpaca}~\footnote{\url{https://huggingface.co/datasets/theblackcat102/evol-codealpaca-v1}}: We sampled 10,000 examples from Evol-Alpaca, which is a code instruction dataset generated by GPT-4 with evol-instruct proposed by WizardCoder~\citep{luo2024wizardcoder}.

\textbf{Python-Code}~\footnote{\url{https://huggingface.co/datasets/ajibawa-2023/Python-Code-23k-ShareGPT}}: We selected 10,000 examples from Python-Code, which comprises instructions and responses generated by GPT-3.5 and GPT-4 for python code generation.

We followed the data processing code in FastChat~\citep{zheng2024judging} to clean instances containing non-English or special characters. 
Then, we split long conversations into blocks with a maximum length of 2048 tokens, resulting in the final \textsc{FuseChat-Mixture} with 95,000 samples.

\vspace{-0.15cm}
\section{Details of Baselines}
\vspace{-0.1cm}
\label{appendix: details_of_baselines}

In this section, we present the details of baseline models compared in our experiments.

\textbf{Proprietary LLMs}: \texttt{GPT-3.5-Turbo-1106}\footnote{\url{https://platform.openai.com/docs/models/gpt-3-5-turbo}}~\citep{achiam2023gpt}, \texttt{Claude-3-Opus}\footnote{\url{https://www.anthropic.com/news/claude-3-family}}~\citep{anthropic2024claude}, and \texttt{GPT-4-1106-Preview}\footnote{\url{https://platform.openai.com/docs/models/gpt-4-turbo-and-gpt-4}}~\citep{achiam2023gpt}. 

\textbf{Source LLMs}: \texttt{OpenChat-3.5-7B}\footnote{\url{https://huggingface.co/openchat/openchat_3.5}}~\citep{wang2024openchat}, \texttt{Starling-LM-7B-alpha}\footnote{\url{https://huggingface.co/berkeley-nest/Starling-LM-7B-alpha}}~\citep{zhu2023starling}, \texttt{NH2-SOLAR-10.7B}\footnote{\url{https://huggingface.co/NousResearch/Nous-Hermes-2-SOLAR-10.7B}}~\citep{kim2023solar}, \texttt{InternLM2-Chat-20B}\footnote{\url{https://huggingface.co/internlm/internlm2-chat-20b}}~\citep{cai2024internlm2}, \texttt{Mixtral-8x7B-Instruct}\footnote{\url{https://huggingface.co/mistralai/Mixtral-8x7B-Instruct-v0.1}}~\citep{jiang2024mixtral}, and \texttt{Qwen-1.5-Chat-72B}\footnote{\url{https://huggingface.co/Qwen/Qwen1.5-72B-Chat}}~\citep{bai2023qwen}. 

\textbf{Ensemble LLMs}: \texttt{Top1-PPL}~\citep{mavromatis2024pack}, which selects the 1st ranked response from source LLMs based on the perplexity of instruction; \texttt{Top1-LLM-Blender}~\citep{jiang2023llm}, which ranks and combines the output text from source LLMs with ranker and fuser models. Due to the fuser model's constraints on maximum sequence length, only the ranker model is utilized to determine and produce the 1st-ranked response; \texttt{Top1-GPT4}~\citep{achiam2023gpt}, which leverages GPT-4 judgment as ranking criteria and yields the 1st ranked response. Since our evaluations also rely on GPT-4, this approach represents an upper bound for comparison.

\textbf{Fused LLMs}: \texttt{OpenChat-3.5-7B SFT}, a special scenario of knowledge fusion with a single source LLM, serves as the supervised fine-tuning baseline using our training dataset; \texttt{OpenChat-3.5-7B Multi} is the knowledge fusion of multiple source chat LLMs simultaneously like \textsc{FuseLLM}~\citep{wan2024knowledge}; \texttt{OpenChat-3.5-7B Starling}, \texttt{OpenChat-3.5-7B SOLAR}, \texttt{OpenChat-3.5-7B InternLM}, \texttt{OpenChat-3.5-7B Mixtral}, and \texttt{OpenChat-3.5-7B Qwen} are target LLMs resulting from pairwise knowledge fusion of the pivot LLM \texttt{OpenChat-3.5-7B} and the rest source LLMs.

\section{Evaluation of Additional Benchmarks}

The primary objective of \textsc{FuseChat} is to integrate multiple chat LLMs into a more powerful model. Consequently, our experiments primarily focus on alignment training data, such as ShareGPT, and chat model evaluation benchmarks like AlpacaEval 2.0 and MT-Bench. In addition to the chat model benchmarks, we also conducted experiments on six general evaluation benchmarks, including MMLU-Pro \citep{wang2024mmlu}, PIQA \citep{bisk2020piqa}, BoolQ \citep{clark2019boolq}, GPQA \citep{rein2023gpqa}, GSM8K \citep{cobbe2021training}, and IFEval \citep{zhou2023instruction}, which assess knowledge understanding, question-answering, common-sense reasoning, and instruction-following. The results are presented in Table \ref{tab:additional_result}. It is important to note that the training data for FuseChat-7B is primarily focused on alignment rather than general knowledge. Therefore, performance improvements on these general benchmarks are less significant compared to those on AlpacaEval 2.0 and MT-Bench. This observation is consistent with recent studies on alignment \citep{meng2024simpo,wu2024self}, which highlight the critical role of alignment dataset construction in determining downstream performance.

\begin{table*}[!ht]
\centering
\resizebox{0.999\linewidth}{!}{
    \begin{tabular}{llllllll}
    \midrule
        \textbf{Model} & \textbf{MMLU-Pro} & \textbf{PIQA} & \textbf{BoolQ} & \textbf{GPQA} & \textbf{GSM8K} & \textbf{IFEval} & \textbf{Average} \\ \midrule
        OpenChat-3.5-7B & 31.63 & 82.86 & 73.91 & 31.30 & 76.88 & 35.73 & 55.38 \\ 
        OpenChat-3.5-7B SFT & 31.32 & 82.75 & 73.91 & 30.30 & 76.04 & 35.34 & 54.94 \\ 
        OpenChat-3.5-7B Multi & 31.39 & 82.43 & 73.73 & 32.30 & 74.75 & 36.25 & 55.14 \\ \midrule
        \textsc{FuseChat}-7B & 31.65(\textcolor{darkblue}{+0.06\%}) & 82.97(\textcolor{darkblue}{+0.13\%}) & 75.50(\textcolor{darkblue}{+2.15\%}) & 37.40(\textcolor{darkblue}{+19.49\%}) & 77.10(\textcolor{darkblue}{+0.29\%}) & 37.49(\textcolor{darkblue}{+4.93\%}) & 57.02(\textcolor{darkblue}{+2.96\%}) \\ \bottomrule 
    \end{tabular}
}
\captionsetup{type=table}
\caption{Comparison results on general evaluation benchmarks.}
\label{tab:additional_result}
\vspace{-0.3cm}
\end{table*}

\section{Case Studies}
\label{appendix: case_studies}

We present case studies to demonstrate the individual strengths of target LLMs (\texttt{OpenChat-3.5-7B Starling}, \texttt{OpenChat-3.5-7B SOLAR}, \texttt{OpenChat-3.5-7B InternLM}, \texttt{OpenChat-3.5-7B Mixtral}, and \texttt{OpenChat-3.5-7B Qwen}) obtained from pairwise knowledge fusion of source LLMs, and show the collective knowledge and strengths of \textsc{FuseChat} (\texttt{\textsc{FuseChat-7B}}) obtained by further merging target LLMs. \texttt{OpenChat-3.5-7B SFT} is used as the baseline for comparison.

\newpage
\begin{table*}[!h]
\centering
\renewcommand\arraystretch{1.1}
\resizebox{0.95\linewidth}{!}{
\begin{tabular}{p{15cm}}
\hline
\textbf{OpenChat-3.5-7B SFT; Math; ID 113;}  \\ \hline
\textbf{Turn 1 Question:} In a survey conducted at a local high school, preferences for a new school color were measured: 58\% of students liked the color blue, 45\% preferred green, and 22\% liked both colors. If we randomly pick a student from the school, what's the probability that they would like neither blue nor green? \\ \hline
\textbf{Turn 1 Answer:}  Let's use the principle of inclusion-exclusion to solve this problem.
The total percentage of students who like either blue or green is the sum of the individual percentages minus the percentage who like both:

\[58\% (blue) + 45\% (green) - 22\% (both) = 75\%\]

This means that 100\% - 75\% = 25\% of students like neither blue nor green. So, the probability that a randomly picked student would not like neither blue nor green is $\boxed{25\%}$. The answer is: 25 \\ \hline
\textbf{Turn 1 Judgment:} The assistant's answer contains a mathematical error in calculating the total percentage of students who like either blue or green. The correct calculation, as shown in the reference answer, should be:

\[58\% + 45\% - 22\% = 81\%\]

This means that 81\% of students like at least one of the colors, not 75\% as the assistant stated. Consequently, the correct percentage of students who like neither color is:

\[100\% - 81\% = 19\%\]

Therefore, the correct probability that a randomly picked student would like neither blue nor green is 19\%, not 25\% as the assistant incorrectly concluded.
The assistant's mistake significantly affects the correctness of the response. However, the assistant did correctly apply the principle of inclusion-exclusion, which indicates an understanding of the correct approach to solving the problem. The error seems to be purely in the arithmetic calculation.
Given the importance of accuracy in mathematical problems and the impact of this error on the final answer, the response's helpfulness is diminished. However, the attempt to apply the correct principle shows some level of understanding. \textcolor{darkblue}{Rating: [[3]]} \\ \hline
\textbf{Turn 2 Question:} If we select a student liked green, what's the probability that he or she would dislike both colors? \\ \hline
\textbf{Turn 2 Answer:} Since 22\% of students like both colors, this means that 78\% of students do not like both colors. Out of these, 45\% prefer green. Therefore, the probability that a student who likes green would dislike the other color is $45\% \times 78\% = 35.7\%$.
So, the probability that a student who likes green would dislike the other color is $\boxed{35.7\%}$. The answer is: 35.7 \\ \hline
\textbf{Turn 2 Judgment:} The assistant's response to the second question contains a fundamental misunderstanding of the question posed. The question asks for the probability that a student who likes green would dislike both colors, which is a contradiction. If a student likes green, they cannot simultaneously dislike green, making it impossible for them to dislike both green and another color (in this case, blue).
The correct answer, as provided in the reference answer, is 0\%. This is because the premise of selecting a student who likes green excludes the possibility of that student disliking both colors. The assistant's calculation of 35.7\% is incorrect because it misinterprets the question as asking for the probability that a student who likes green would dislike the other color (blue), which was not the question asked.
Therefore, the assistant's response is incorrect due to a misunderstanding of the question's premise, leading to an erroneous calculation that does not apply to the scenario described.
 \textcolor{darkblue}{Rating: [[1]]} \\ \hline
\end{tabular}
}
\caption{Case studies on MT-Bench. The answers are generated by \texttt{OpenChat-3.5-7B SFT}.}
\label{tab:case-1-clm}
\end{table*}

\newpage
\begin{table*}[!h]
\centering
\renewcommand\arraystretch{1.1}
\resizebox{0.95\linewidth}{!}{
\begin{tabular}{p{15cm}}
\hline
\textbf{OpenChat-3.5-7B Starling; Math; ID 113;}  \\ \hline
\textbf{Turn 1 Question:} In a survey conducted at a local high school, preferences for a new school color were measured: 58\% of students liked the color blue, 45\% preferred green, and 22\% liked both colors. If we randomly pick a student from the school, what's the probability that they would like neither blue nor green? \\ \hline
\textbf{Turn 1 Answer:} We can solve this problem using the principle of inclusion-exclusion.
The total number of students who like either blue or green is the sum of those who like blue, those who like green, minus those who like both (to avoid double-counting).
So, the number of students who like either color is:

\[58\% (blue) + 45\% (green) - 22\% (both) = 75\%\]

This means that 25\% of students like neither color.
Therefore, the probability that a randomly picked student likes neither color is $\boxed{25\%}$. The answer is: 25 \\ \hline
\textbf{Turn 1 Judgment:} The assistant's answer contains a mathematical error in calculating the percentage of students who like either blue or green. The correct calculation, as shown in the reference answer, should be:

\[ \text{At least one of the colors} = 58\% + 45\% - 22\% = 103\% - 22\% = 81\% \]

This means that 81\% of students like at least one of the colors, not 75\% as the assistant stated. Consequently, the percentage of students who like neither color is not 25\% but rather:

\[ \text{Neither color} = 100\% - 81\% = 19\% \]

Therefore, the correct probability that a randomly picked student likes neither color is 19\%, not 25\%.
The assistant's mistake significantly affects the correctness of the response. The correct approach and calculation were provided in the reference answer, which the assistant failed to replicate accurately. This error directly impacts the helpfulness of the response, as it provides the user with incorrect information.
Given the importance of accuracy in mathematical problems and the potential confusion caused by incorrect percentages, the assistant's response is rated lower due to the critical error in calculation. \textcolor{darkblue}{Rating: [[3]]} \\ \hline
\textbf{Turn 2 Question:} If we select a student liked green, what's the probability that he or she would dislike both colors? \\ \hline
\textbf{Turn 2 Answer:} Since the student dislikes both colors, they must dislike green. Therefore, the probability that a student who likes green would dislike both colors is 0\%.  \\ \hline
\textbf{Turn 2 Judgment:} The assistant's response to the second question is correct and aligns with the reference answer. It correctly states that if a student likes green, the probability that they would dislike both colors (blue and green) is 0\%. This is because the premise of the question already establishes that the student likes green, making it impossible for them to dislike both colors simultaneously. The explanation is clear, concise, and logically sound, directly addressing the user's question without unnecessary complexity.
The assistant's answer to the second question does not contain any mistakes and is both correct and helpful. It provides a straightforward explanation that directly answers the user's query, making it easy for the user to understand the concept being discussed. \textcolor{darkblue}{Rating: [[10]]} \\ \hline
\end{tabular}
}
\caption{Case studies on MT-Bench. The answers are generated by \texttt{OpenChat-3.5-7B Starling}.}
\label{tab:case-1-starling}
\end{table*}

\newpage
\begin{table*}[!h]
\centering
\renewcommand\arraystretch{1.1}
\resizebox{0.95\linewidth}{!}{
\begin{tabular}{p{15cm}}
\hline
\textbf{OpenChat-3.5-7B SOLAR; Math; ID 113;}  \\ \hline
\textbf{Turn 1 Question:} In a survey conducted at a local high school, preferences for a new school color were measured: 58\% of students liked the color blue, 45\% preferred green, and 22\% liked both colors. If we randomly pick a student from the school, what's the probability that they would like neither blue nor green? \\ \hline
\textbf{Turn 1 Answer:} We can solve this problem using the principle of inclusion-exclusion.
The total number of students who like either blue or green is the sum of those who like blue and those who like green, minus the number who like both. That is:

\[\text{Total} = \text{Blue} + \text{Green} - \text{Both}\]

We are given that 58\% of students like blue, 45\% like green, and 22\% like both. So, we can calculate the percentage of students who like neither as follows:

\[\text{Neither} = \text{Total} - \text{Blue} - \text{Green} + \text{Both}\]

Substituting the given percentages:

\[\text{Neither} = 100\% - (58\% + 45\% - 22\%)\]

Simplifying:

\[\text{Neither} = 100\% - 75\%\]

Therefore, the probability that a randomly picked student would like neither blue nor green is 25\%. \\ \hline
\textbf{Turn 1 Judgment:} The assistant's answer contains a critical error in the calculation process. The mistake lies in the incorrect application of the principle of inclusion-exclusion and the subsequent calculation of the percentage of students who like neither color.
The correct application of the principle of inclusion-exclusion for calculating the percentage of students who like at least one of the colors (blue or green) is as follows:

\[ \text{At least one of the colors} = (\text{Blue}) + (\text{Green}) - (\text{Both}) \]

\[ \text{At least one of the colors} = 58\% + 45\% - 22\% = 103\% - 22\% = 81\% \]

This means 81\% of students like at least one of the colors, blue or green. To find the percentage of students who like neither color, we subtract this percentage from 100\%:

\[ \text{Neither color} = 100\% - \text{At least one of the colors} \]

\[ \text{Neither color} = 100\% - 81\% = 19\% \]

Therefore, the correct probability that a randomly picked student would like neither blue nor green is 19\%, not 25\% as the assistant incorrectly calculated.
The assistant's mistake was in the formula used for calculating the percentage of students who like neither color. The assistant incorrectly attempted to subtract the sum of individual preferences minus the overlap from 100\%, which does not align with the principle of inclusion-exclusion correctly applied in this context.
Given the significant error in the calculation that led to an incorrect final answer, the response's correctness is compromised. However, the assistant did correctly identify the need to use the principle of inclusion-exclusion, which shows an understanding of the correct approach, albeit misapplied.  \textcolor{darkblue}{Rating: [[3]]} \\ \hline
\textbf{Turn 2 Question:} If we select a student liked green, what's the probability that he or she would dislike both colors? \\ \hline
\textbf{Turn 2 Answer:} If a student is selected who likes green, the probability that he or she would dislike both colors is 1 - the probability that they like both colors.
We know that 22\% of students like both colors. So, the probability that a student who likes green dislikes both colors is:

\[1 - 22\% = 78\%\]

Therefore, the probability that a student who likes green would dislike both colors is 78\%. \\ \hline
\textbf{Turn 2 Judgment:} The assistant's response to the second question contains a fundamental misunderstanding of the scenario presented. According to the reference answer, if a student likes green, the probability that they would dislike both colors is 0\%. This is logically consistent because liking green precludes the possibility of disliking both green and blue simultaneously. The assistant's answer, suggesting a 78\% probability that a student who likes green would dislike both colors, is incorrect. The correct answer should acknowledge that once a student is known to like green, they cannot be in the category of disliking both colors, making the probability 0\%.
 \textcolor{darkblue}{Rating: [[1]]} \\ \hline
\end{tabular}
}
\caption{Case studies on MT-Bench. The answers are generated by \texttt{OpenChat-3.5-7B SOLAR}.}
\label{tab:case-1-solar}
\end{table*}

\newpage
\begin{table*}[!h]
\centering
\renewcommand\arraystretch{1.1}
\resizebox{0.95\linewidth}{!}{
\begin{tabular}{p{15cm}}
\hline
\textbf{OpenChat-3.5-7B InternLM; Math; ID 113;}  \\ \hline
\textbf{Turn 1 Question:} In a survey conducted at a local high school, preferences for a new school color were measured: 58\% of students liked the color blue, 45\% preferred green, and 22\% liked both colors. If we randomly pick a student from the school, what's the probability that they would like neither blue nor green? \\ \hline
\textbf{Turn 1 Answer:}  To find the probability that a student likes neither blue nor green, we need to subtract the probability that they like at least one of the colors from 1.
Let A be the event that a student likes blue, and B be the event that a student likes green. We are given that $P(A) = 0.58$, $P(B) = 0.45$, and $P(A \cap B) = 0.22$.
First, we need to find the probability that a student likes at least one of the colors. We can use the formula for the probability of the union of two events:

\[P(A \cup B) = P(A) + P(B) - P(A \cap B)\]
\[P(A \cup B) = 0.58 + 0.45 - 0.22 = 0.75\]

Now, to find the probability that a student likes neither blue nor green, we subtract the probability that they like at least one of the colors from 1:

\[P(neither) = 1 - P(at least one color)\]
\[P(neither) = 1 - 0.75 = 0.25\]

So, the probability that a randomly picked student would like neither blue nor green is 0.25 or 25\%. \\ \hline
\textbf{Turn 1 Judgment:} The assistant's answer contains a mathematical error in calculating the probability that a student likes at least one of the colors. The assistant incorrectly calculates the probability of a student liking at least one of the colors (blue or green) as 0.75 (or 75\%), which leads to an incorrect final answer for the probability of a student liking neither color as 0.25 (or 25\%).
The correct calculation, as shown in the reference answer, should be as follows:

- The probability of liking blue (P(Blue)) = 58\% or 0.58

- The probability of preferring green (P(Green)) = 45\% or 0.45

- The probability of liking both colors (P(Both)) = 22\% or 0.22

Using the principle of inclusion-exclusion:
\[ \text{At least one of the colors} = P(Blue) + P(Green) - P(Both) = 0.58 + 0.45 - 0.22 = 0.81 \]

This means 81\% of students like at least one of the colors, and thus:
\[ \text{Neither color} = 1 - \text{At least one of the colors} = 1 - 0.81 = 0.19 \]

Therefore, the correct probability that a randomly picked student would like neither blue nor green is 0.19 or 19\%, not 25\% as the assistant's answer suggests.
The assistant's mistake lies in the incorrect calculation of the union of the two events, which led to an inaccurate conclusion. This error significantly impacts the correctness of the response.
Given the importance of accuracy in probability calculations and the impact of this error on the final answer, the assistant's response is rated lower for its correctness. However, the assistant did correctly apply the principle of inclusion-exclusion and provided a clear explanation of the steps, which is helpful for understanding the process. \textcolor{darkblue}{Rating: [[4]]} \\ \hline
\textbf{Turn 2 Question:} If we select a student liked green, what's the probability that he or she would dislike both colors? \\ \hline
\textbf{Turn 2 Answer:} The question seems to be asking for the probability that a student who likes green would dislike both colors, which is not possible because the student already likes green. The student cannot dislike both colors while also liking one of them.
 \\ \hline
\textbf{Turn 2 Judgment:} The assistant's response to the second question is correct and aligns with the reference answer. It correctly identifies the logical inconsistency in the question: if a student likes green, they cannot simultaneously dislike both green and another color. The assistant succinctly explains why the probability that a student who likes green would dislike both colors is 0\%, which is the same conclusion reached in the reference answer. This explanation is clear, logically sound, and directly addresses the user's question, demonstrating an understanding of basic probability principles and the specific scenario presented.
  \textcolor{darkblue}{Rating: [[10]]} \\ \hline
\end{tabular}
}
\caption{Case studies on MT-Bench. The answers are generated by \texttt{OpenChat-3.5-7B InternLM}.}
\label{tab:case-1-internlm}
\end{table*}

\newpage
\begin{table*}[!h]
\centering
\renewcommand\arraystretch{1.1}
\resizebox{0.95\linewidth}{!}{
\begin{tabular}{p{15cm}}
\hline
\textbf{OpenChat-3.5-7B Mixtral; Math; ID 113;}  \\ \hline
\textbf{Turn 1 Question:} In a survey conducted at a local high school, preferences for a new school color were measured: 58\% of students liked the color blue, 45\% preferred green, and 22\% liked both colors. If we randomly pick a student from the school, what's the probability that they would like neither blue nor green? \\ \hline
\textbf{Turn 1 Answer:} The probability that a student likes blue is 58\%, and the probability that they like green is 45\%. Since 22\% of students like both colors, we can subtract this from the total percentage to find the percentage of students who like either blue or green but not both.

\[58\% + 45\% - 22\% = 71\%\]

So, 71\% of students like either blue or green, which means 100\% - 71\% = 29\% of students like neither color.
Therefore, the probability that a randomly picked student would like neither blue nor green is $\boxed{29\%}$. The answer is: 29 \\ \hline
\textbf{Turn 1 Judgment:} The assistant's answer contains a critical mathematical error in calculating the probability of students liking neither blue nor green. The correct approach, as outlined in the reference answer, involves using the principle of inclusion-exclusion to accurately account for the overlap between students who like both colors.
Correct Calculation:

- The correct calculation to find the percentage of students who like at least one of the colors (blue or green) is to add the percentages of students who like each color and then subtract the percentage of students who like both (to correct for double counting), which is \(58\% + 45\% - 22\% = 81\%\).

- Therefore, to find the percentage of students who like neither color, we subtract the percentage of students who like at least one of the colors from 100\%, resulting in \(100\% - 81\% = 19\%\).

The assistant incorrectly calculated the percentage of students who like either blue or green but not both as 71\%, leading to the incorrect conclusion that 29\% of students like neither color. The correct probability that a randomly picked student would like neither blue nor green is 19\%, not 29\%.
Given the significant mathematical error and the resulting incorrect conclusion, the assistant's answer is not helpful for someone trying to understand how to correctly apply the principle of inclusion-exclusion in probability. \textcolor{darkblue}{Rating: [[2]]} \\ \hline
\textbf{Turn 2 Question:} If we select a student liked green, what's the probability that he or she would dislike both colors? \\ \hline
\textbf{Turn 2 Answer:} If a student is selected who likes green, the probability that they dislike both colors is equal to the percentage of students who like both colors.
In this case, 22\% of students like both colors, so the probability that a student who likes green would dislike both colors is $\boxed{22\%}$.The answer is: 22
 \\ \hline
\textbf{Turn 2 Judgment:}  The assistant's response to the second question contains a significant error. The correct answer, as outlined in the reference answer, is that the probability of a student who likes green disliking both colors is 0\%. This is because the premise of the question already establishes that the student likes green, making it impossible for them to dislike both green and blue simultaneously. The assistant's answer incorrectly states that the probability is 22\%, which seems to be a misunderstanding of the question's premise.
The correct approach should have recognized the logical contradiction in the question: a student who likes green cannot at the same time dislike both green and another color. Therefore, the probability of a student who likes green also disliking both colors is 0\%.
Given this significant mistake and misunderstanding of the question's premise, the response is incorrect and misleading. \textcolor{darkblue}{Rating: [[1]]} \\ \hline
\end{tabular}
}
\caption{Case studies on MT-Bench. The answers are generated by \texttt{OpenChat-3.5-7B Mixtral}.}
\label{tab:case-1-mixtral}
\end{table*}

\newpage
\begin{table*}[!h]
\centering
\renewcommand\arraystretch{1.1}
\resizebox{0.95\linewidth}{!}{
\begin{tabular}{p{15cm}}
\hline
\textbf{OpenChat-3.5-7B Qwen; Math; ID 113;}  \\ \hline
\textbf{Turn 1 Question:} In a survey conducted at a local high school, preferences for a new school color were measured: 58\% of students liked the color blue, 45\% preferred green, and 22\% liked both colors. If we randomly pick a student from the school, what's the probability that they would like neither blue nor green? \\ \hline
\textbf{Turn 1 Answer:} To find the probability that a student likes neither blue nor green, we can use the formula for the probability of the union of two events:

\[P(A \cup B) = P(A) + P(B) - P(A \cap B)\]

In this case, A represents the event that a student likes blue, and B represents the event that a student likes green. We are given that $P(A) = 0.58$, $P(B) = 0.45$, and $P(A \cap B) = 0.22$.
First, we need to find the probability of the event that a student likes either blue or green, which is represented by $P(A \cup B)$:

\[P(A \cup B) = P(A) + P(B) - P(A \cap B) = 0.58 + 0.45 - 0.22 = 0.75\]

Now, we need to find the probability of the event that a student likes neither blue nor green, which is represented by $P(A' \cap B')$:

\[P(A' \cap B') = 1 - P(A \cup B) = 1 - 0.75 = 0.25\]

Therefore, the probability that a randomly picked student would like neither blue nor green is 0.25 or 25\%. \\ \hline
\textbf{Turn 1 Judgment:} The assistant's answer contains a mathematical error in calculating the probability of a student liking either blue or green. The assistant incorrectly calculates the sum of the probabilities as 0.75 (or 75\%) instead of correctly calculating it as 81\%, as shown in the reference answer. This error leads to an incorrect final answer for the probability of a student liking neither blue nor green.
Correct Calculation based on the reference answer:

- The correct calculation for the probability of a student liking at least one of the colors (blue or green) is 58\% + 45\% - 22\% = 81\%.

- Therefore, the correct probability of a student liking neither color is 100\% - 81\% = 19\%.

The assistant's answer incorrectly concludes that the probability is 25\%, whereas the correct probability, as per the reference answer and the correct calculation, is 19\%.
Despite the assistant's method being fundamentally sound and the explanation clear, the mathematical mistake significantly impacts the correctness of the final answer. The assistant's approach correctly applies the principle of inclusion-exclusion but fails in the execution of basic arithmetic.
Given the importance of accuracy in mathematical problems and the impact of this error on the final answer, the assistant's response must be rated lower. However, the explanation and approach were correctly aligned with the principle of inclusion-exclusion, which is a positive aspect of the response. \textcolor{darkblue}{Rating: [[4]]} \\ \hline
\textbf{Turn 2 Question:} If we select a student liked green, what's the probability that he or she would dislike both colors? \\ \hline
\textbf{Turn 2 Answer:} If a student is selected who likes green, we want to find the probability that they dislike both colors. Since the student likes green, they cannot dislike both colors. Therefore, the probability that a student who likes green would dislike both colors is 0.
  \\ \hline
\textbf{Turn 2 Judgment:} The assistant's response to the second question is correct and aligns with the reference answer. It accurately explains that if a student is already known to like green, it is impossible for them to dislike both green and blue simultaneously. Therefore, the probability that a student who likes green would dislike both colors is indeed 0\%. This explanation is clear, concise, and directly addresses the user's question without unnecessary complexity. The assistant's answer is helpful in clarifying the logical impossibility of the scenario proposed by the user, thus preventing any confusion about the nature of probability in this context.
 \textcolor{darkblue}{Rating: [[10]]} \\ \hline
\end{tabular}
}
\caption{Case studies on MT-Bench. The answers are generated by \texttt{OpenChat-3.5-7B Qwen}.}
\label{tab:case-1-qwen}
\end{table*}

\newpage
\begin{table*}[!h]
\centering
\renewcommand\arraystretch{1.1}
\resizebox{0.95\linewidth}{!}{
\begin{tabular}{p{15cm}}
\hline
\textbf{\textsc{FuseChat-7B}; Math; ID 113;}  \\ \hline
\textbf{Turn 1 Question:} In a survey conducted at a local high school, preferences for a new school color were measured: 58\% of students liked the color blue, 45\% preferred green, and 22\% liked both colors. If we randomly pick a student from the school, what's the probability that they would like neither blue nor green? \\ \hline
\textbf{Turn 1 Answer:} To find the probability that a student likes neither blue nor green, we can use the formula:

\[ P(\text{neither}) = 1 - P(\text{blue}) - P(\text{green}) + P(\text{both}) \]

where P(blue) is the probability that a student likes blue, P(green) is the probability that a student likes green, and P(both) is the probability that a student likes both colors.
We are given that $P(\text{blue}) = 58\%$, $P(\text{green}) = 45\%$, and $P(\text{both}) = 22\%$. Plugging these values into the formula, we get:

\[ P(\text{neither}) = 1 - 0.58 - 0.45 + 0.22 \]
\[ P(\text{neither}) = 1 - 0.85 + 0.22 \]
\[ P(\text{neither}) = 0.15 \]

So, the probability that a randomly picked student would like neither blue nor green is 15\%.  \\ \hline
\textbf{Turn 1 Judgment:} The assistant's answer uses the principle of inclusion-exclusion correctly but makes a calculation error in the final step. According to the assistant's formula:

\[ P(\text{neither}) = 1 - P(\text{blue}) - P(\text{green}) + P(\text{both}) \]

The values given are:
- \( P(\text{blue}) = 58\% = 0.58 \)
- \( P(\text{green}) = 45\% = 0.45 \)
- \( P(\text{both}) = 22\% = 0.22 \)

Plugging these into the formula, the assistant calculates:

\[ P(\text{neither}) = 1 - 0.58 - 0.45 + 0.22 \]

However, the assistant incorrectly calculates the final result as 0.15 (or 15\%). The correct calculation should be:

\[ P(\text{neither}) = 1 - 0.58 - 0.45 + 0.22 = 1 - 1.03 + 0.22 = 1 - 0.81 = 0.19 \]

Therefore, the correct probability that a randomly picked student would like neither blue nor green is 19\%, not 15\%.
The reference answer correctly calculates this probability as 19\% using the principle of inclusion-exclusion and correctly accounts for the overlap between students who like both colors. The assistant's approach is fundamentally correct but falters at the arithmetic stage, leading to an incorrect final answer.
Given the mistake in calculation, the assistant's answer is not entirely correct, but since the method used was appropriate, it shows a good understanding of the principle involved. Therefore, the assistant's response, while flawed in execution, is still somewhat helpful for understanding how to approach the problem. \textcolor{darkblue}{Rating: [[6]]} \\ \hline
\textbf{Turn 2 Question:} If we select a student liked green, what's the probability that he or she would dislike both colors? \\ \hline
\textbf{Turn 2 Answer:} If a student is selected who likes green, we want to find the probability that they dislike both colors. Since the student likes green, they cannot dislike both colors. Therefore, the probability that a student who likes green would dislike both colors is 0\%.
\\ \hline
\textbf{Turn 2 Judgment:} The assistant's answer to the second question is correct and aligns with the reference answer. It accurately explains that if a student likes green, the probability of them disliking both colors (blue and green) is 0\%. This is because the premise of the question already establishes that the student likes green, making it impossible for them to dislike both colors simultaneously. The explanation is clear, concise, and logically sound, effectively addressing the user's query without any errors or unnecessary complexity. \textcolor{darkblue}{Rating: [[10]]} \\ \hline
\end{tabular}
}
\caption{Case studies on MT-Bench. The answers are generated by \texttt{\textsc{FuseChat-7B}}.}
\label{tab:case-1-fusechat}
\end{table*}

\newpage

\end{document}